\documentclass[letterpaper, 10 pt, conference]{ieeeconf}  % Comment this line out if you need a4paper

\IEEEoverridecommandlockouts                              % This command is only needed if 
\usepackage{graphicx}
\usepackage{subcaption} % For subfigures and subcaptions
\usepackage{xcolor}
\usepackage{multirow}
\usepackage{hyperref}
\usepackage{url}
\usepackage{cite}

\title{\LARGE \bf Analyzing Behaviors of Mixed Traffic via Reinforcement Learning at Unsignalized  Intersections}

\author{Supriya Sarker \\ \texttt{ssarker8@vols.utk.edu}}

\begin{document}

\maketitle
\thispagestyle{empty}
\pagestyle{empty}

%%%%%%%%%%%%%%%%%%%%%%%%%%%%%%%%%%%%%%%%%%%%%%%%%%%%%%%%%%%%%%%%%%%%%%%%%%%%%%%%

\begin{abstract}

In this report, we delve into two critical research inquiries. Firstly, we explore the extent to which Reinforcement Learning (RL) agents exhibit multimodal distributions in the context of stop-and-go traffic scenarios. Secondly, we investigate how RL-controlled Robot Vehicles (RVs) effectively navigate their direction and coordinate with other vehicles in complex traffic environments. Our analysis encompasses an examination of multimodality within queue length, outflow, and platoon size distributions for both Robot and Human-driven Vehicles (HVs). Additionally, we assess the Pearson coefficient correlation, shedding light on relationships between queue length and outflow, considering both identical and differing travel directions. Furthermore, we delve into causal inference models, shedding light on the factors influencing queue length across scenarios involving varying travel directions. Through these investigations, this report contributes valuable insights into the behaviors of mixed traffic (RVs and HVs) in traffic management and coordination.

\end{abstract}

\section{INTRODUCTION} 
We are now entering an exciting era of fully autonomous vehicles. Companies like Waymo \cite{waymo} have already deployed fully automated vehicles in vibrant urban centers such as Phoenix and San Francisco, and they have ambitious plans to extend their presence to additional cities in Washington and California~\cite{waymo2022citybybay}. Similarly, Cruise \cite{cruise} is making strides in this field, operating in key locations like Austin, San Francisco, and Phoenix~\cite{cruise2023onemillion}. The operation of autonomous vehicles in urban environments has spurred significant interest among intelligent transportation researchers. Urban traffic complexity arises from the diverse mix of vehicles, each following its own set of rules, signs, signals, and negotiation styles. Moreover, these regulations can vary significantly from state to state and country to country, adding to the intricacy of the traffic landscape. 

Navigating through urban areas can be quite demanding, especially when it comes to dealing with intersections. In 2021, the National Highway Traffic Safety Administration (NHTSA)~\cite{nhtsa} reported a concerning statistic. Out of the 10.8 million crashes that occurred in the United States, resulting in fatalities, injuries, or property damage, a significant 4.9 million (approximately 45\%) unfolded at intersections. This not only poses a serious safety concern but also translates into substantial economic burdens, encompassing costs associated with waiting times, collisions, and the energy expended while waiting.  

Traditional methods of controlling intersections include traffic lights and stop signs. Traffic lights, which have been in use since the early 20th century, work by allowing for platooning, where vehicles in the permitted direction can pass without negotiating the right-of-way with others, simply following the leader. However, the phase of platooning is fixed in a particular direction. The fixed phase platooning may create a longer queue and waiting time in a certain direction of intersection.

To minimize delays, such as waiting times, numerous transportation researchers have put forth a variety of methods. Among these approaches, learning environments using reinforcement learning (RL) agents have emerged as particularly effective solutions~\cite{villarreal2023can,villarreal2023hybrid,wang2023learning,Wang2023Privacy}. Certain endeavors have encountered challenges due to their assumption of 100\% Connected and Autonomous Vehicles (CAV) in traffic scenarios. It's important to acknowledge that as we move toward a future with full autonomy, there will be a transitional phase of mixed autonomy where human-driven vehicles (HVs) and  robot vehicles (RVs) coexist on the roads. Over the past few years, RVs have emerged as a promising solution for enhancing traffic performance. When compared to HVs, RVs have the ability to perceive a broader range of factors and respond with remarkable speed. This holds great promise for elevating both the efficiency and safety of urban traffic. 

Centralized mixed traffic control systems have notable disadvantages. Firstly, they are vulnerable to single points of failure, such as technical glitches or cyberattacks, which can disrupt the entire traffic management system. Secondly, these systems necessitate extensive data collection and processing from all vehicles, raising concerns about data privacy and surveillance. Additionally, centralized control relies on continuous communication between vehicles and the central authority, leading to communication overhead and potential latency issues. Scaling up such systems to accommodate a growing number of vehicles can be complex and costly, limiting their adaptability in resource-constrained areas. Lastly, centralized control often demands substantial infrastructure upgrades and investments, making it less flexible in regions with limited resources. These limitations underscore the importance of carefully considering the trade-offs when choosing between centralized and decentralized approaches in mixed traffic scenarios. 

In response to the notable demerits associated with centralized mixed traffic control, transportation researchers are increasingly shifting their focus toward decentralized mixed autonomy research. This transition is driven by the recognition of several key benefits that decentralized systems offer. Decentralized mixed traffic control provides greater scalability, enabling easy adaptation to varying traffic conditions. It reduces vulnerability by mitigating risks associated with single points of failure or cyberattacks. Real-time decision-making capabilities at the local level lead to faster responses, while the preservation of data privacy addresses concerns about surveillance and data security. Moreover, decentralized control enhances traffic flow, promoting smoother interactions on the road by allowing vehicles to adapt autonomously to localized conditions. These advantages underscore the growing interest and potential of decentralized approaches in advancing mixed traffic management and autonomous vehicle technologies. Nevertheless, the analysis of how a decentralized mixed autonomous system functions as a traffic light has yet to be explored. Within our research, we've delved into several decentralized approaches. Among these, we've chosen to focus on the recent work by Wang et al.~\cite{wang2023learning} as our primary reference for this analysis, as it introduces an innovative decentralized mixed unsignalized traffic control system. 

Hence, the driving force behind this report is to delve into the intricacies of RVs within the context of unsignalized intersection control in mixed traffic. To kickstart our exploration, we posit the hypothesis that ``Reinforcement Learning (RL) functions akin to a miniature traffic signal.'' In our quest to substantiate this hypothesis, we have formulated a set of precise research inquiries that we aim to address conclusively by the report's conclusion. These research questions are as follows:

\begin{itemize}
    \item RQ1: To what extent does the distribution of RL agents become multimodal during a stop-and-go scenario?
    \item RQ2: How does the RV effectively manage its own direction and coordinate with others?
\end{itemize}

Therefore, the main contributions made in this report are 
\begin{itemize}
    \item we investigate the multimodality of queue length, outflow, and platoon size distribution of RV and HV;
    \item we analyze the pearson coefficient correlation between the queue length and outflow of the same and two different directions at a time; and 
    \item we investigate the causal inference of queue length the same and two different directions at a time.
\end{itemize}   

The rest of the paper is organized as follows. The state-of-the-art centralized and decentralized methods of controlled unsignalized intersections are discussed in Section~\ref{related}. The background of the framework is discussed in Section~\ref{background}. The data processing of the report is discussed in Section~\ref{data_processing}. Section~\ref{data_analysis} presents the analysis of the reference paper. Section~\ref{conclusion} draws conclusions.

\section{RELATED WORK}
\label{related}
There have been both centralized and decentralized approaches to control intersection traffic. 
\subsection{Centralized Mixed Autonomy Intersection Traffic Management}
Centralized mixed autonomy intersection traffic refers to a traffic management system where the control and coordination of vehicles, both autonomous and human-driven, at an intersection or multiple intersections are overseen and directed by a central authority or traffic management system. In this setup, decisions related to traffic signal timing, lane assignments, and the prioritization of vehicles are made at a centralized location and then communicated to vehicles within the intersection area. Centralized mixed autonomy intersection traffic aims to optimize traffic flow, enhance safety, and reduce congestion through centralized control and coordination of various types of vehicles. 

Based on the assumption that in the future most vehicles are autonomous and connected, Dresner and Stone proposed a novel intersection control protocol denoted Autonomous Intersection Management (AIM) \cite{dresner2008multiagent}. AIM protocol consists of two types of autonomous agents - intersection managers and driver agents. Intersection managers are responsible for directing the vehicles through the intersections, while the driver agents are responsible for controlling the CAV to which they are assigned. To improve the throughput and efficiency of the system, the driver agents call ahead to the intersection manager and request a path reservation. The intersection manager then determines whether or not this request can be met by checking whether it conflicts
with any previously approved reservation or a potential HV. Depending on the approval of the intersection manager. Connected and autonomous vehicles (CAVs), with the help of advanced sensing devices, are more accurate and predictable compared to human-operated vehicles (HVs) \cite{fajardo2011automated}. On the other hand, AIM, assuming 100\% of the vehicles are CAVs, was shown to reduce the
delay imposed on vehicles by orders of magnitude compared to traffic signals. Moreover, AIM did not perform better than traffic signals when more than 10\% of the vehicles are HVs. 

To prove AIM, Sharon and Stone \cite{sharon2017protocol} proposed a Hybrid Autonomous Intersection Management (H-AIM) to overcome the inefficiency of the AIM in the large portion of the transition period. Unlike AIM, H-AIM assumes sensing of approaching vehicles which allows it to identify approaching HVs. If no HV is observed on a given lane, then trajectories originating from that lane are assumed to not be occupied by HVs, allowing AVs more flexibility in obtaining reservations. A single lane entering a four-way intersection can allow three different turning possibilities (turn left, continue straight, turn right) or any combination of the three. H-AIM grants reservation in an FCFS order. 

Deep Reinforcement Learning (DRL) has gained popularity in understanding real-road traffic situations and improving the performance of autonomous vehicles due to its ability to handle complex and dynamic environments. Unlike traditional rule-based systems that rely on predefined instructions, DRL allows autonomous vehicles to learn from experience and adapt to real-world scenarios. By leveraging neural networks and continuous learning, DRL enables vehicles to make data-driven decisions in real time, which is crucial for navigating the unpredictability of traffic conditions. Using deep reinforcement learning, Jang et.al \cite {jang2019simulation} trained a set of two autonomous vehicles to lead a fleet of vehicles onto a roundabout and then transferred two policies -(1) a policy with noise injected into the state and action space and (2) a policy without any injected noise from simulation to a scaled city without fine-tuning. The noise-injected policy consistently performs an acceptable metering behavior, implying that the noise eventually aids with the zero-shot policy transfer.  

To control connected vehicles in the vicinity of a signal-free intersection and minimize the total vehicle delay Yang and Oguchi~\cite{yang2020intelligent} proposed a model to predict the overall delay of one intersection with the information collected by connected vehicles followed by the development of an advanced vehicle control system based on the delay prediction to achieve the goal of minimizing intersection delay. 

\subsection{Decentralized Mixed Autonomy Intersection Traffic Management}
Decentralized Mixed Autonomy Intersection Traffic refers to a traffic management system where the control and coordination of vehicles, including autonomous and human-driven ones, at an intersection or multiple intersections are primarily conducted by individual vehicles or local systems. In this setup, vehicles have a higher degree of autonomy and are equipped with the capability to make real-time decisions based on their local perception of the traffic environment. This decentralized approach often involves vehicles communicating with each other and with the surrounding infrastructure to negotiate right-of-way and navigate through intersections safely. Unlike centralized control systems, decentralized mixed autonomy intersection traffic relies on distributed decision-making, where each vehicle plays an active role in adapting to and managing traffic conditions at the intersection. In decentralized mixed autonomy, vehicles have a higher degree of autonomy in making local decisions based on real-time information. 

%\textcolor{red}{\textbf{Describe Dawei's work here}}
Wang et al. \cite{wang2023learning} proposed a decentralized RL approach to handle current challenges. They introduced an innovative decentralized reinforcement learning (RL) approach to effectively address these challenges. Their method involved defining a control zone around the intersection, providing a comprehensive view of the traffic conditions within this zone. Once a vehicle entered this designated area, their proposed method took control.

Within the control zone, each robot vehicle (RV) skillfully encoded its perception of traffic conditions into a concise, fixed-length representation. This representation encompassed traffic data from eight distinct directions, capturing both macroscopic aspects like queue length and waiting times, as well as microscopic details such as the precise locations of vehicles within the intersection. 

Subsequently, each RV positioned in front of the intersection entrance line utilized this representation to make informed high-level decisions, determining whether to 'Stop' or 'Go.' Importantly, these high-level decisions from various RVs at the entrance line were thoughtfully shared and coordinated through vehicle-to-vehicle (V2V) communication.

Finally, guided by their high-level decisions, each RV seamlessly traversed the intersection, employing a precise low-level control mechanism to ensure safe and efficient passage.

\section{Background Study}
\label{background}
\subsection{Intersection Geometry}
Intersection geometries can vary significantly based on the number of roads or legs converging at a junction. In this section we will discuss various kinds of intersection geometries, including 3-leg intersections and 4-leg intersections, with some examples and context. 
\subsubsection{3-Leg Intersection}
A 3-leg intersection, often known as a T-intersection, is a fundamental configuration where three roadways meet at a single point. It typically comprises a primary road, or the "stem" of the T, intersecting with two smaller roads, forming the "arms" of the T. T-intersections can be found in urban and rural settings alike, often featuring traffic control measures such as stop signs or traffic signals to regulate vehicle flow. In urban areas, proper signage and pedestrian accommodations are crucial for safety, while in rural regions, drivers on the main road usually have the right-of-way.

\subsubsection{4-Leg Intersection}
A 4-leg intersection, also referred to as a four-way intersection, represents one of the most common intersection geometries. It involves four roadways converging at a central point, creating a square or rectangular layout. To manage traffic, four-way intersections often employ traffic signals, stop signs, or roundabouts. These intersections can vary in complexity, ranging from simple residential intersections to intricate urban intersections with multiple lanes, pedestrian crossings, and advanced signal systems. They are prevalent in urban areas and serve as crucial junctions for transportation networks.

\subsubsection{Roundabout (Circular Intersection)}
A roundabout is a circular intersection where traffic circulates counterclockwise around a central island. Unlike traditional intersections with stop signs or traffic signals, roundabouts rely on yield signs at entry points, fostering a continuous flow of traffic. They are celebrated for their safety benefits, as they mitigate the severity of collisions and minimize the need for abrupt stops. Roundabouts can be found in both urban and suburban settings and come in various sizes to accommodate varying traffic volumes and vehicle types.

\subsubsection{Diamond Interchange (Cloverleaf Interchange)}
A diamond interchange connects two major roads, often highways or expressways, using entrance and exit ramps arranged in a cloverleaf or diamond-shaped pattern. This design aims to facilitate the smooth flow of traffic between intersecting roadways, often incorporating overpasses or underpasses to separate merging and exiting traffic. Diamond interchanges are prevalent on highways and expressways, providing efficient access to different routes while maintaining the continuous flow of traffic on the main roadways.

\subsubsection{Diverging Diamond Interchange)}
A Diverging Diamond Interchange (DDI) is a modern interchange design that briefly shifts traffic to the opposite side of the road within the interchange. This temporary realignment enables more efficient left-turn movements and helps alleviate congestion. DDIs are gaining popularity in highway design due to their capacity to enhance traffic flow and safety, particularly in locations with high traffic volumes and complex interchange configurations.

\subsubsection{Continuous Flow Intersection}
A Continuous Flow Intersection (CFI) is tailored to enhance left-turn movements at intersections. It integrates specialized lanes that allow left-turning traffic to proceed before reaching the primary intersection, reducing traffic conflicts and congestion. CFIs are often deployed in areas with substantial traffic loads to optimize overall traffic flow and operational efficiency, offering a solution to the challenges posed by traditional intersection layouts.

\subsection{Traffic Light in Intersection}
The classification of intersections based on traffic signals is a fundamental way to understand how traffic is controlled and managed at these critical points in the transportation network. Intersections can be classified based on the availability of traffic signals into three primary categories: signalized intersections, unsignalized intersections, and roundabouts. 
\subsubsection{Signalized Intersections}
Signalized intersections are intersections equipped with traffic signals (also known as traffic lights) to control the flow of vehicles and pedestrians. Traffic signals regulate the right-of-way for different directions of traffic, alternating between green (go), yellow (prepare to stop), and red (stop) signals. They are effective at managing traffic flow and reducing accidents. 
\subsubsection{Unsignalized Intersections}
Unsignalized intersections are intersections without traffic signals, where the right-of-way is typically determined by yield signs, stop signs, or general right-of-way rules. Vehicles must yield or stop as required by signage and applicable traffic laws. Right-of-way is often determined by who arrives first or other established rules. 
\subsubsection{Roundabouts}
Roundabouts are circular intersections designed to facilitate continuous traffic flow without the use of traffic signals or stop signs. Vehicles in a roundabout yield to circulating traffic and enter when a safe gap is available. Traffic flows counterclockwise around a central island. they promote safer and more efficient traffic flow.

\subsection{Mixed Traffic Control}
Mixed traffic represents a dynamic and evolving transportation scenario where autonomous or robot vehicles coexist alongside traditional human-driven vehicles within the same interconnected road network. In this context, envision the autonomous vehicle as an independent decision-maker, with its surroundings, consisting of the road layout and traffic conditions, forming the canvas upon which it navigates. The guiding light for this autonomous entity is a reward signal, typically encapsulating measures of driving excellence, safety, and operational efficiency.

To effectively employ Reinforcement Learning (RL), our autonomous vehicle must first create an accurate and comprehensive representation of its state. This representation encompasses crucial details such as the vehicle's current location, speed, orientation, the presence of nearby objects, the status of traffic signals, and more. This state representation serves as the foundational input to our RL model, setting the stage for informed decision-making.

Within this context, the action space comes into play, defining the spectrum of choices available to our autonomous vehicle. These actions encompass pivotal decisions such as acceleration, braking, steering maneuvers, and lane changes. The RL agent, operating within its current state, orchestrates its actions with the overarching goal of maximizing cumulative rewards over time.

The rewards themselves are the linchpin of this learning process, acting as evaluators of performance. They assign numerical values to states and actions, guiding the vehicle toward behaviors aligned with driving safety, compliance with traffic rules, and optimal navigation. For instance, rewards can be structured around concepts like maintaining a safe following distance, adhering to speed limits, and avoiding collisions.

To accomplish this formidable task, RL models like Q-learning or Deep Q-Networks (DQN) step in, refining their understanding of the environment by iteratively estimating the value of specific actions in various states. These models empower the RL agent to explore different courses of action, observing the outcomes and adjusting its strategies accordingly.

During training, the autonomous vehicle actively engages with a simulated environment or controlled real-world scenarios. It employs the RL model to select actions guided by its current state and learns from the resulting rewards. Special exploration strategies are employed to encourage the vehicle to experiment with diverse actions and glean insights into their consequences.

As this journey unfolds, the RL agent accumulates experiences and continuously refines its policy to optimize the expected cumulative reward. The RL model stands as a steadfast ally, aiding the vehicle in adapting its decisions to the ever-changing dynamics of traffic conditions and in fine-tuning its driving behavior.

In this quest for autonomous driving prowess, developers often turn to simulations, which provide a safe and controlled playground for training and evaluating RL-based policies. These simulations allow vehicles to traverse a spectrum of scenarios, from routine to rare and complex, without exposure to real-world risks, fostering a resilient and adaptive autonomous driving ecosystem.

\subsection{Mixed intersection management}
Within our report, we recognize the importance of making complex terminology accessible and comprehensible to our readers. To ensure clarity and understanding, we have thoughtfully identified and defined key terms and concepts used throughout the document.  
\begin{itemize}
  \item \textbf{Approaching the intersection}
  Approaching the intersection refers to the count of vehicles that have entered the control zone and are proceeding toward the intersection from a specific direction. These vehicles are actively moving closer to the intersection and are within the designated control area as they approach the junction. 
  \item \textbf{Outflow}
  Outflow refers to the vehicles that have entered the intersection but have not yet completed their passage through it. These are vehicles that are currently within the intersection, in the process of navigating it, but have not yet exited to continue in their intended direction.

  \item \textbf{Exit}
  Exit refers to the count of vehicles that have effectively traversed the intersection and continued on their path in a different direction. To illustrate, consider a scenario where a vehicle initially embarks on a journey in the 'left-straight' direction. As it makes its way from the left side of the intersection, successfully navigating through the junction node, and ultimately progressing in a 'straight to right' direction, it becomes part of the outflow category.  
  \item \textbf{Queue:} Formed when vehicles are moving at a speed slower than $0.1$m/s, in the same lane. Sometimes due to reaction time, they can be a bit far.

  \item \textbf{Waiting time}
  Waiting time refers to the amount of time that a vehicle spends stationary or in a state of reduced movement while waiting for a specific event or condition to change.

\end{itemize}

\section{DATA PROCESSING}
\label{data_processing}

%\textcolor{red}{\textbf{This section must answer- How to get the data?}}

\subsection{Intersections Architecture}

In our referenced study by Wang et al. \cite{wang2023learning}, we embarked on a comprehensive examination of traffic dynamics at four distinct junctions: junctions 229, 332, 334, and 499. These junctions, while sharing the commonality of being 4-leg intersections, each present a unique configuration with a varying number of lanes.

To facilitate clarity and precision in our analysis, Wang introduced a nomenclature that encompassed eight distinct directional categories: 'top-straight,' 'top-left,' 'right-straight,' 'right-left,' 'bottom-straight,' 'bottom-left,' 'left-straight,' and 'left-left.' These designations capture the diverse ways in which vehicles approach and exit these intersections, painting a comprehensive picture of traffic flow patterns.

Moreover, to ensure ease of identification and traceability, each lane within every directional category at each junction was thoughtfully assigned a specific lane number. This meticulous lane numbering system simplifies the process of pinpointing and interpreting data collected from specific lanes within a given directional category at any of the four junctions.

\subsection{Data acquisition and pre-processing}
%From the trajectory data, we have extracted timestep, queue length, wait time, the action of the head vehicle of a lane, traffic light phase of a particular lane, outflow, approach, exit, lane-wise wait for each lane of every direction of four junctions. The queue length of all lanes of the same direction of a junction and the wait time of all lanes of the same direction of a junction are later aggregated so that when we use queue length and wait time data we only consider the queue length and wait time of each direction disregarding specific lane. However, for lane-wise wait time, we consider waiting time in a particular lane in the direction of a junction.  

%%%%%%% By ChatGPT (Need revision again)
From our comprehensive trajectory dataset, we've meticulously extracted a rich array of traffic insights. These include valuable information such as the temporal progression (timestep), the extent of congestion (queue length), the duration of waits (wait time), the decisive actions taken by the head vehicle in each lane, the current traffic light phase in specific lanes, outflow rates, approach patterns, exit points, and the lane-specific wait times. These details have been meticulously collected for every lane within each direction across four distinct junctions.

To streamline the data for practical analysis, we employ a thoughtful aggregation process. Specifically, we consolidate the queue lengths for all lanes within the same direction at a given junction. Likewise, we aggregate the wait times for all lanes in a particular direction at each junction. This approach allows us to simplify our analysis, focusing on the overall queue lengths and wait times for each directional flow while disregarding the intricate specifics of individual lanes.

However, when it comes to lane-wise wait times, we continue to consider the wait times associated with each specific lane within a particular directional flow at each junction. This distinction ensures that we retain detailed insights into lane-specific performance, which can be vital for addressing specific traffic challenges and optimizing lane-level strategies.

% \subsubsection{Model Selection}
% \textcolor{red}{\textbf{Bibek's explanation will be here and so explanation of \ref{fig_model_1} and \ref{fig_model_2} will be here}}

% \begin{figure*}
%   \includegraphics[width=\textwidth]{fig/model_selection/model_selection 1.jpg}
%   \caption{Model Selection 1(Dummy title)}
%   \label{fig_model_1}
% \end{figure*}

% \begin{figure*}
%   \includegraphics[width=\textwidth]{fig/model_selection/model_selection 2.jpg}
%   \caption{Model Selection 2(Dummy title)}
%   \label{fig_model_2}
% \end{figure*}

\section{Data Analysis}
\label{data_analysis}
In the data analysis section, we delve into several key analytical metrics for both RV and HV. These metrics provide valuable insights into the system's performance. We examine the Pearson Coefficient Correlation (PCC) between queue length and outflow, explore the frequency distribution of queue length and outflow, and investigate the frequency distribution of platoon size. Each of these metrics offers a unique perspective on the behavior of the traffic system.

\subsubsection{Pearson Coefficient Correlation}
This is a statistic that quantifies the degree to which two variables are linearly related. It measures the strength and direction (positive or negative) of the linear relationship between two continuous variables. It's a valuable tool in statistics and data analysis to assess the degree of association between variables. \\
\begin{equation}
r = \frac{\sum{(X_i - \bar{X})(Y_i - \bar{Y})}}{\sqrt{\sum{(X_i - \bar{X})^2} \sum{(Y_i - \bar{Y})^2}}}
\label{eq:pearson}
\end{equation}
The Pearson correlation coefficient, denoted as "r," ranges from -1 to 1. If r is close to 1, it indicates a strong positive linear relationship. If r is close to -1, it indicates a strong negative linear relationship. If r is close to 0, it indicates a weak or no linear relationship. 

The motivation behind conducting Pearson Coefficient Correlation (PCC) analysis is to delve into the intricate interplay of queue lengths in different directions. It allows us to unearth patterns and relationships within the data, particularly highlighting instances where queue lengths in specific directions move in tandem. Positive PCC values between two directional queue lengths signify a synchronization in their fluctuations, whether it's an increase or decrease.

\begin{figure*}[htbp]
\centering
\begin{subfigure}{0.25\textwidth}
   \includegraphics[width=\linewidth]{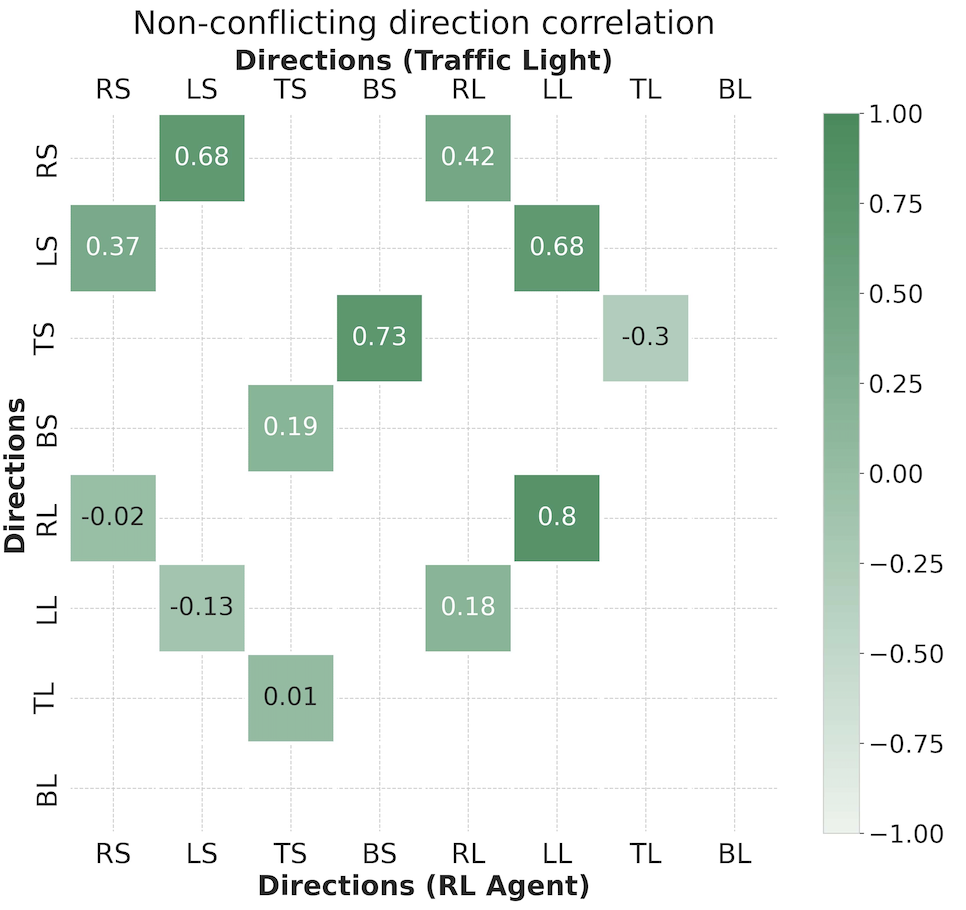}
   \caption{}
   \label{fig:ql_non_conflict_229}
\end{subfigure}%
\begin{subfigure}{0.25\textwidth}
   \includegraphics[width=\linewidth]{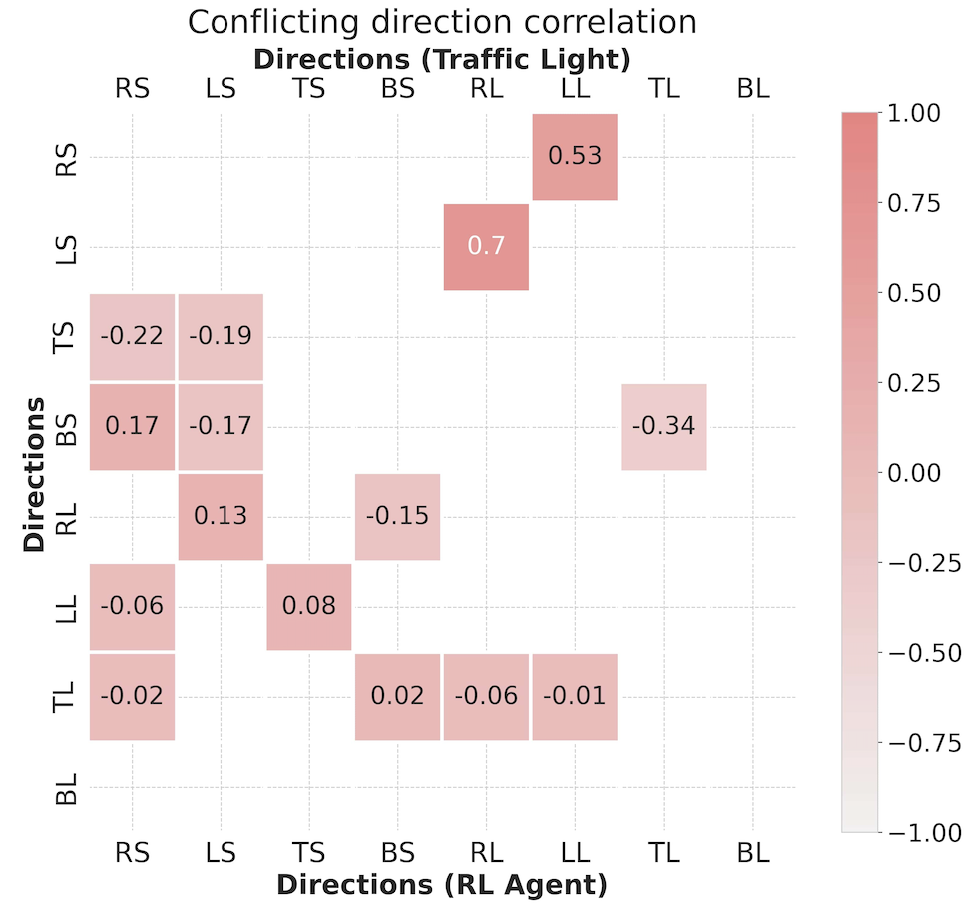}
   \caption{}
   \label{fig:ql_conflict_229}
\end{subfigure}%
\begin{subfigure}{0.25\textwidth}
   \includegraphics[width=\linewidth]{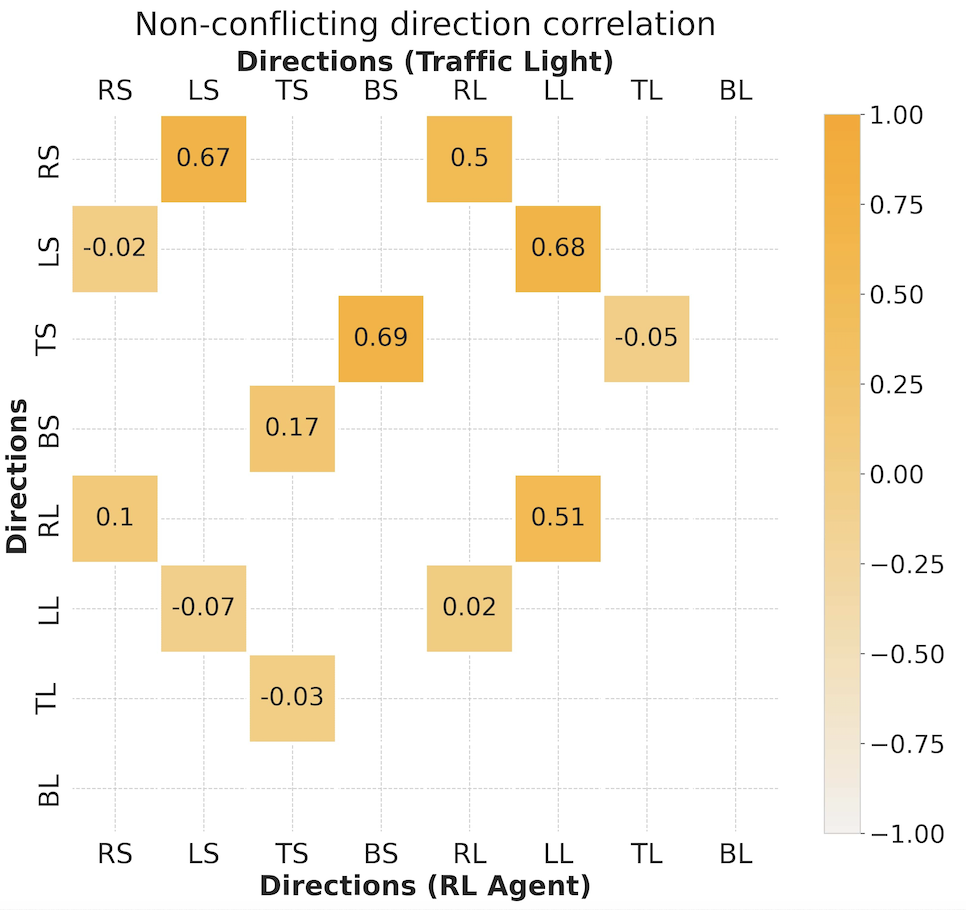}
   \caption{}
   \label{fig:outflow_non_conflict_229}
\end{subfigure}%
\begin{subfigure}{0.25\textwidth}
   \includegraphics[width=\linewidth]{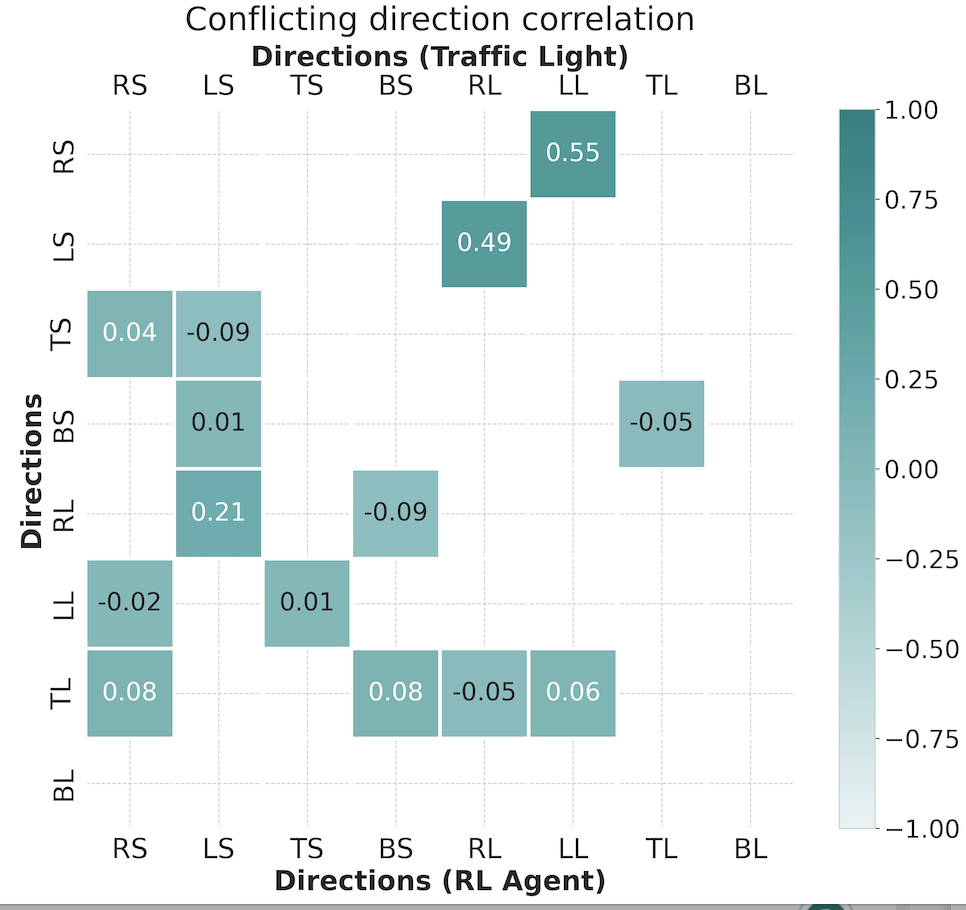}
   \caption{}
   \label{fig:outflow_conflict_229}
\end{subfigure}

\caption{PCC of queue length in (a) non-conflicting directions (b) conflicting directions for junction 299, outflow in (c) non-conflicting directions (d) conflicting directions for junction 299}
\label{fig:pcc_ql_outflow_229}
\end{figure*}

In Fig~\ref{fig:ql_non_conflict_229} and ~\ref{fig:ql_conflict_229}  depicted PCC of queue length in non-conflicting and conflicting directions, respectively and in fig.~\ref{fig:outflow_non_conflict_229} and ~\ref{fig:outflow_conflict_229} the outflow of non-conflicting and conflicting directions, respectively of junction 229 for both RL agents (diagonally bottom half) and TL (diagonally upper half). For all the junctions one phenomenon is significantly noticeable that is the PCC (either positive or negative) exits for RL compared to TL, especially for conflicting directions. On the other hand, there is less variation in TL which results in missing correlation in several directions. It turns out that for non-conflicting directions PCC value is higher than conflicting directions. For TL conflicting directions PCC reflected that there is no such correlation which means the traffic light mechanism maintains a fixed traffic light phase that is programmed in it disregarding the particular necessity of the traffic scenarios. 
On the other hand, RL can understand the traffic scenarios and tries to synchronize the conflicting directions to avoid congestion and longer queues. This analysis slightly answers to our RQ2 which is how RL organizes directions. Also, it supports our hypothesis that RL works as a mini-traffic light.   
%Answer the RQ

%%%%%%%%%%%%%%%%%%%%%%%%%%%%%%%%%%%%%%%%%%%%%%%%%%%%
\begin{figure*}[htbp]
\centering
\begin{subfigure}{0.25\textwidth}
   \includegraphics[width=\linewidth]{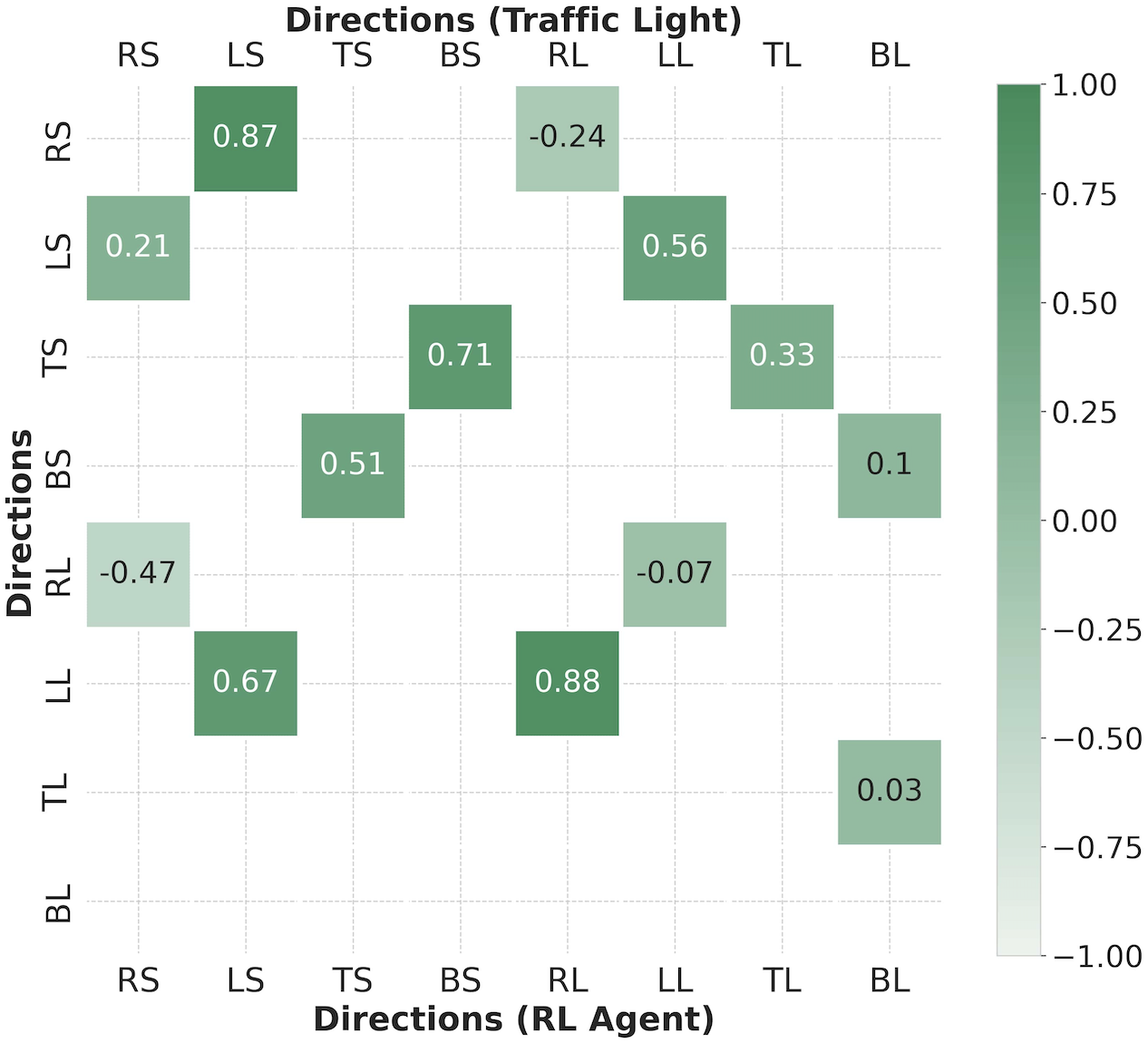}
   \caption{}
   \label{fig:ql_non_conflict_332}
\end{subfigure}%
\begin{subfigure}{0.25\textwidth}
   \includegraphics[width=\linewidth]{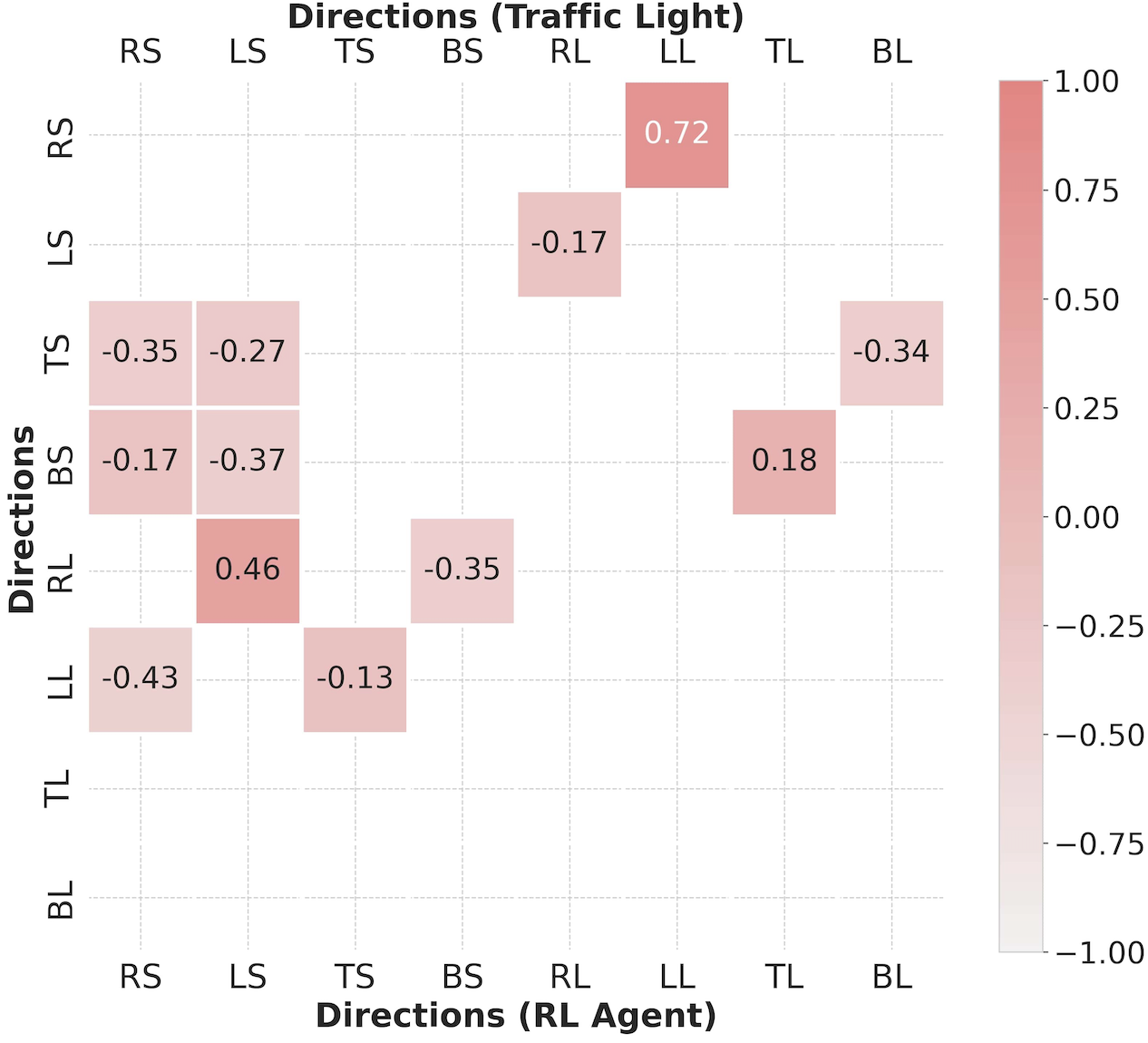}
   \caption{}
   \label{fig:ql_conflict_332}
\end{subfigure}%
\begin{subfigure}{0.25\textwidth}
   \includegraphics[width=\linewidth]{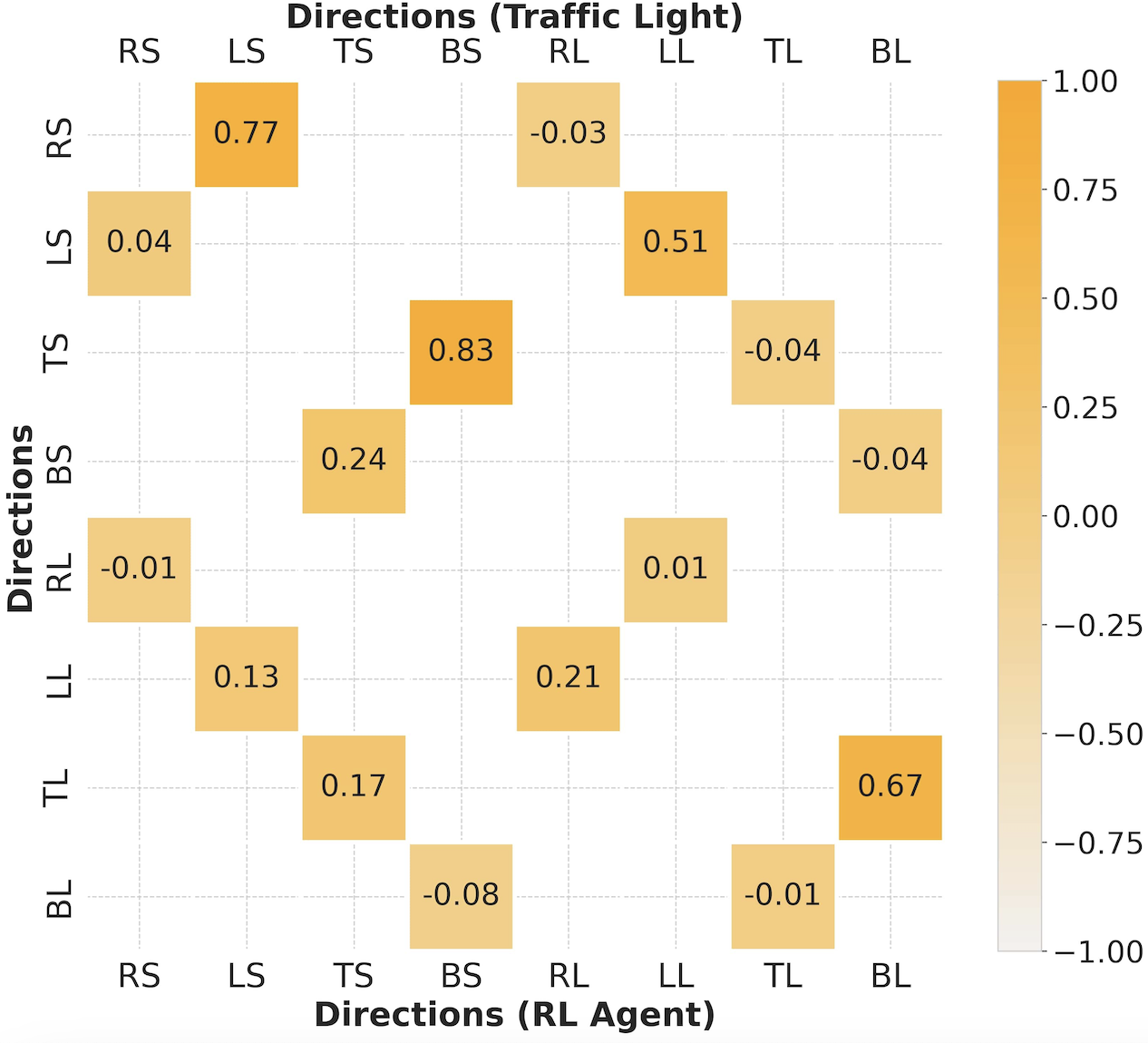}
   \caption{}
   \label{fig:outflow_non_conflict_332}
\end{subfigure}%
\begin{subfigure}{0.25\textwidth}
   \includegraphics[width=\linewidth]{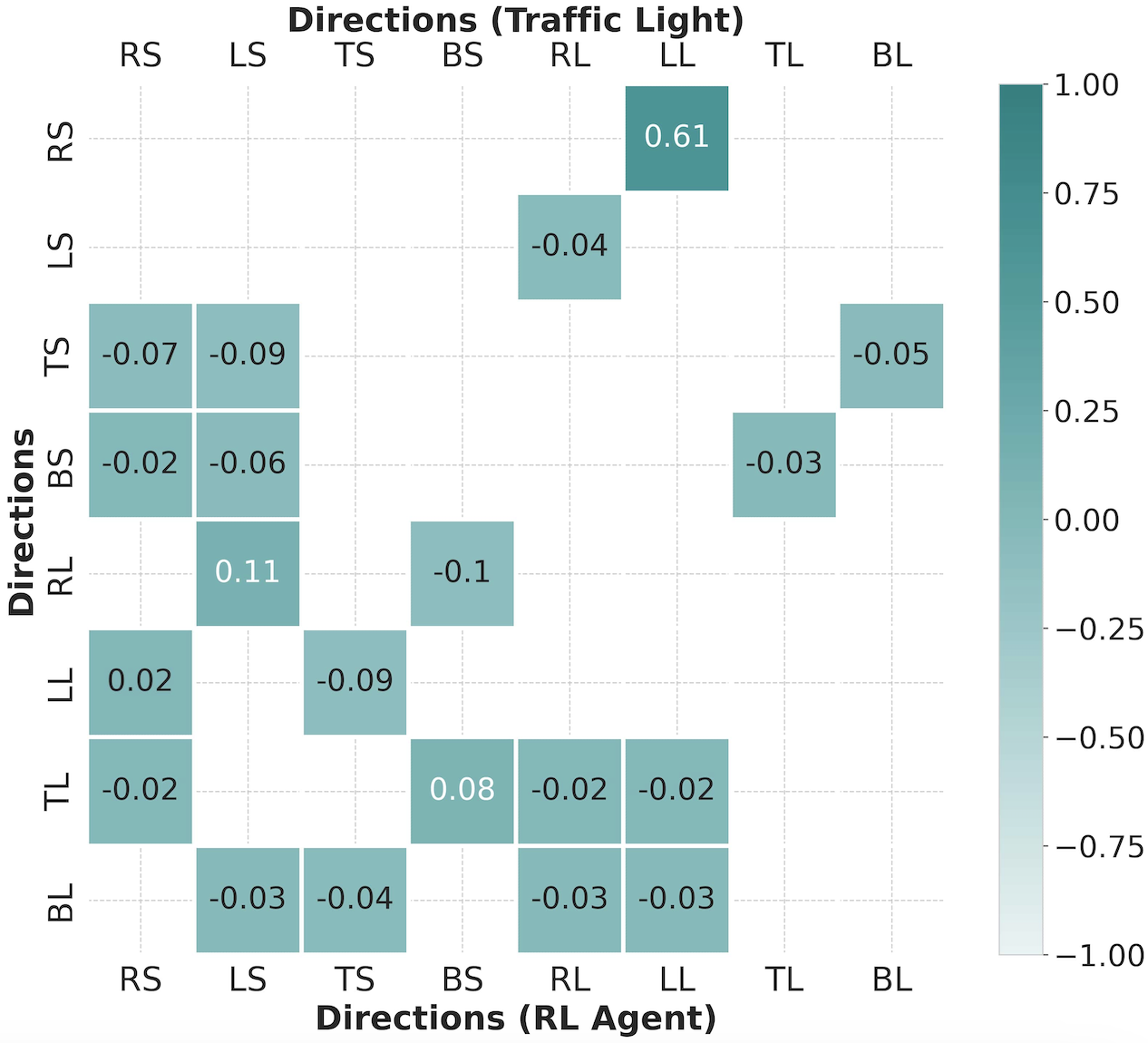}
   \caption{}
   \label{fig:outflow_conflict_332}
\end{subfigure}

\caption{PCC of queue length in (a) non-conflicting directions (b) conflicting directions for junction 332, outflow in (c) non-conflicting directions (d) conflicting directions for junction 332}
\label{fig:pcc_ql_outflow_332}
\end{figure*}

%%%%%%%%%%%%%%%%%%%%%%%%%%%%%%%%%%%%%%%%%%%%%%%%%%%%
\begin{figure*}[htbp]
\centering
\begin{subfigure}{0.25\textwidth}
   \includegraphics[width=\linewidth]{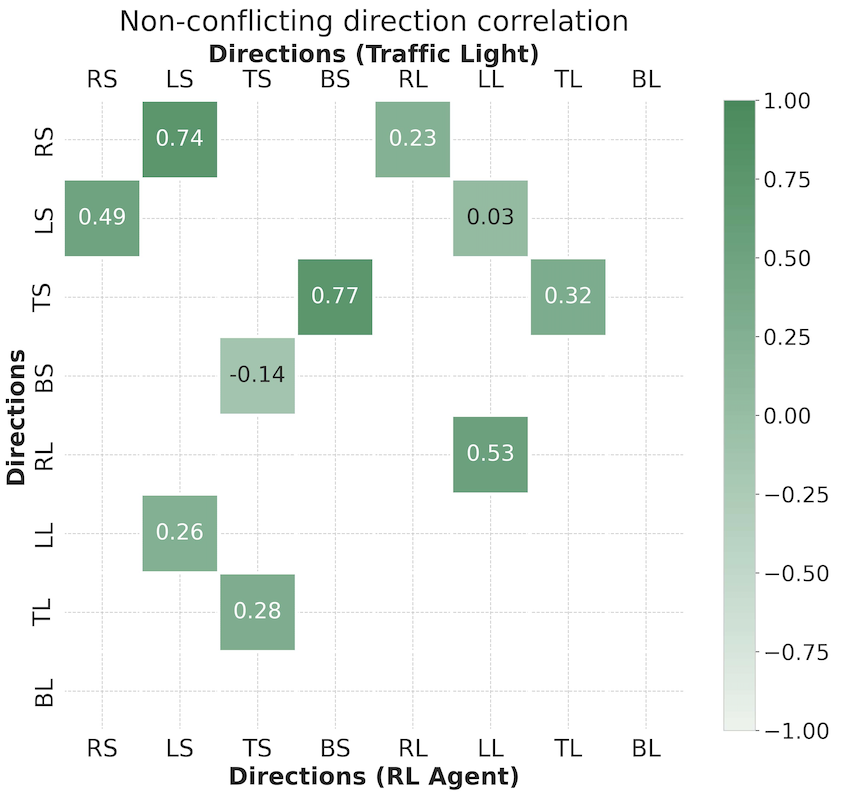}
   \caption{}
   \label{fig:ql_non_conflict_334}
\end{subfigure}%
\begin{subfigure}{0.25\textwidth}
   \includegraphics[width=\linewidth]{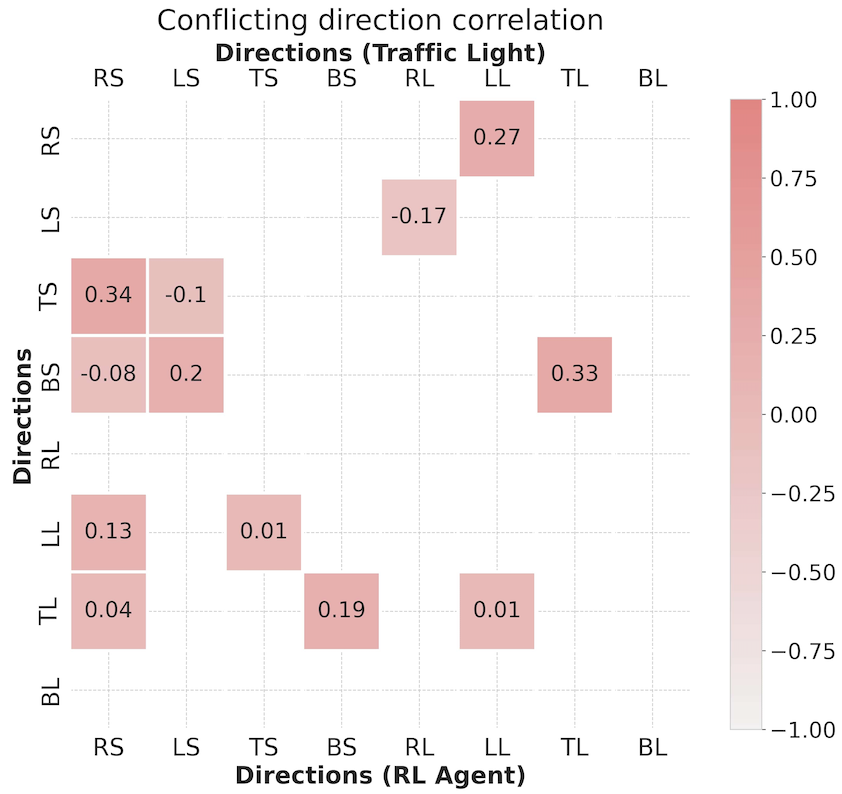}
   \caption{}
   \label{fig:ql_conflict_334}
\end{subfigure}%
\begin{subfigure}{0.25\textwidth}
   \includegraphics[width=\linewidth]{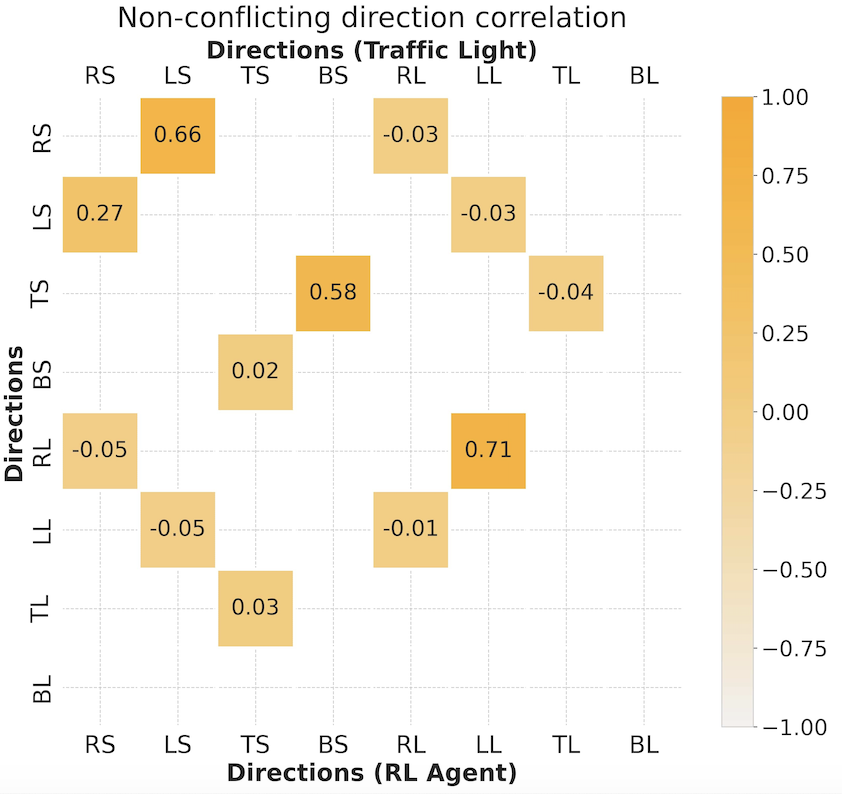}
   \caption{}
   \label{fig:outflow_non_conflict_334}
\end{subfigure}%
\begin{subfigure}{0.25\textwidth}
   \includegraphics[width=\linewidth]{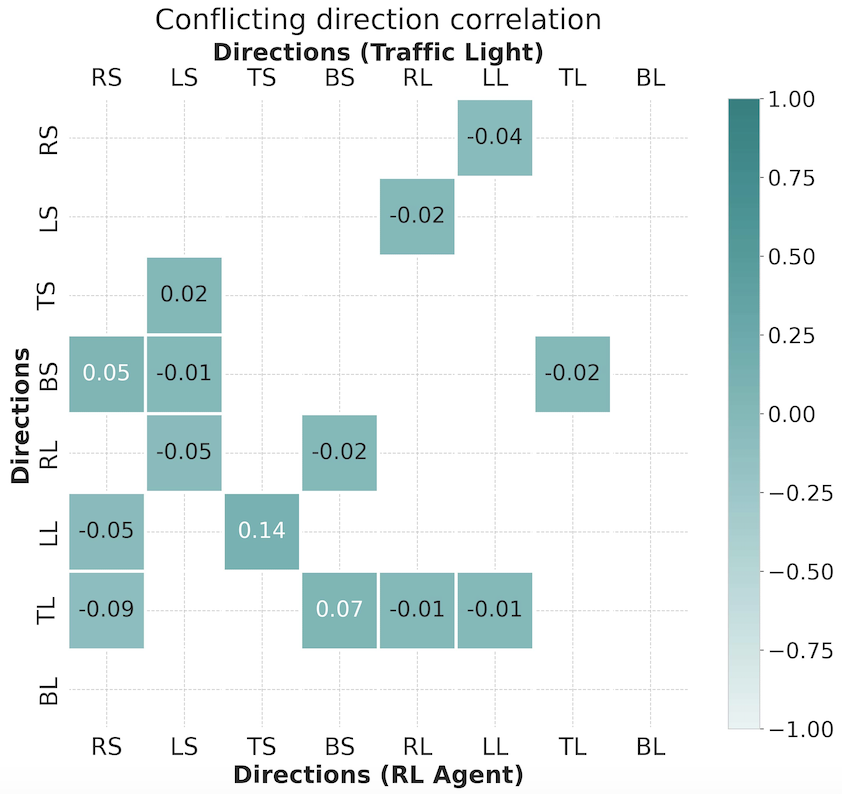}
   \caption{}
   \label{fig:outflow_conflict_334}
\end{subfigure}

\caption{PCC of queue length in (a) non-conflicting directions (b) conflicting directions for junction 334, outflow in (c) non-conflicting directions (d) conflicting directions for junction 334}
\label{fig:pcc_ql_outflow_334}
\end{figure*}
%%%%%%%%%%%%%%%%%%%%%%%%%%%%
\begin{figure*}[htbp]
\centering
\begin{subfigure}{0.25\textwidth}
   \includegraphics[width=\linewidth]{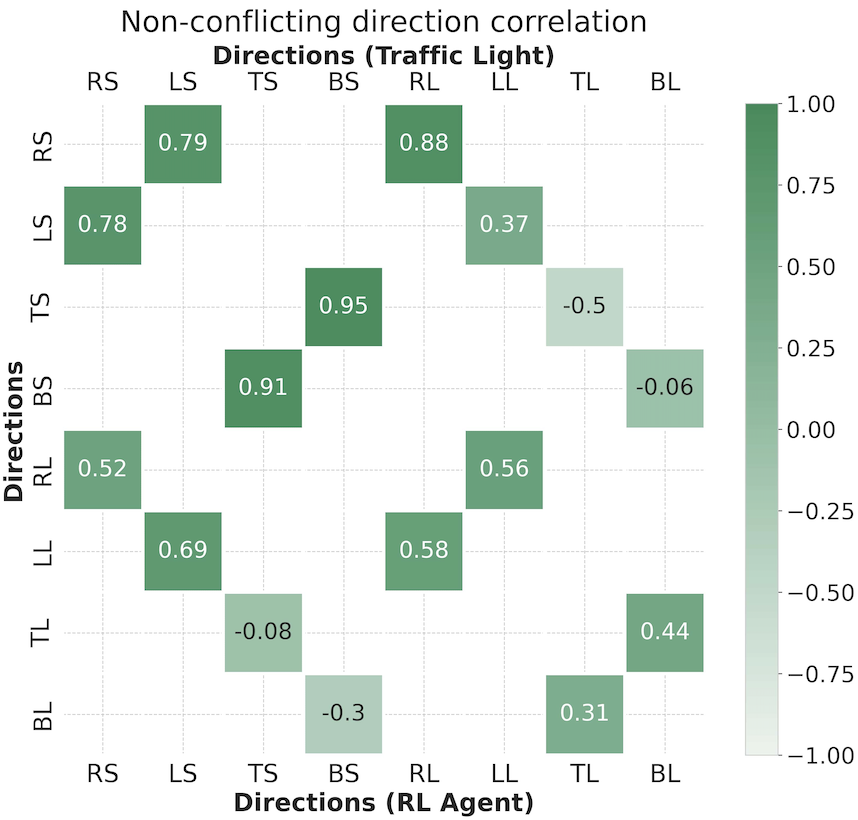}
   \caption{}
   \label{fig:ql_non_conflict_499}
\end{subfigure}%
\begin{subfigure}{0.25\textwidth}
   \includegraphics[width=\linewidth]{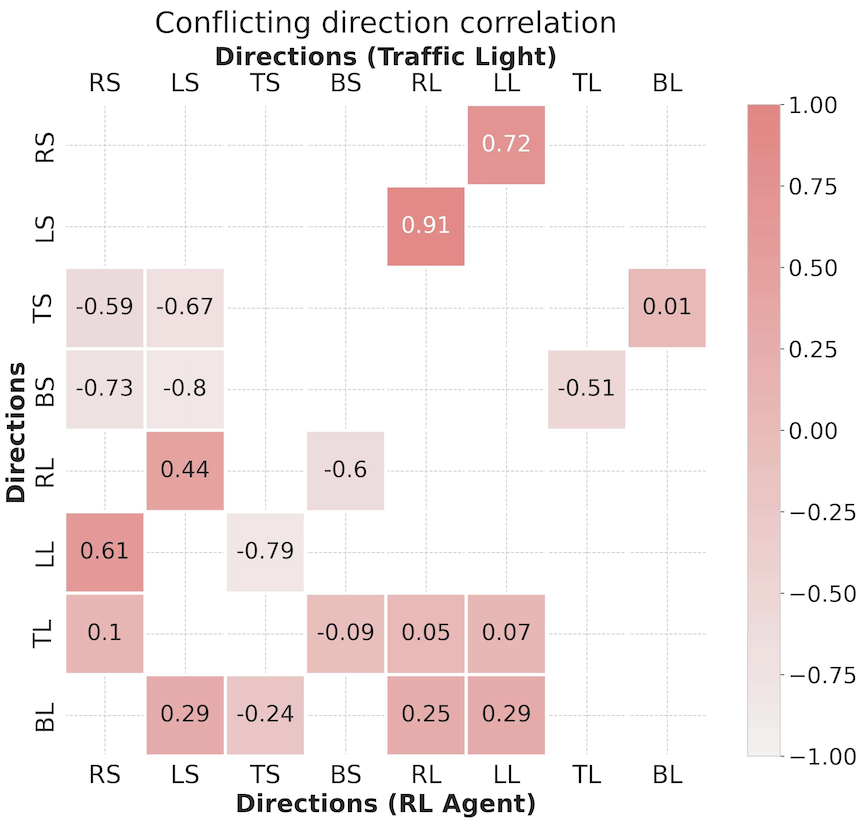}
   \caption{}
   \label{fig:ql_conflict_499}
\end{subfigure}%
\begin{subfigure}{0.25\textwidth}
   \includegraphics[width=\linewidth]{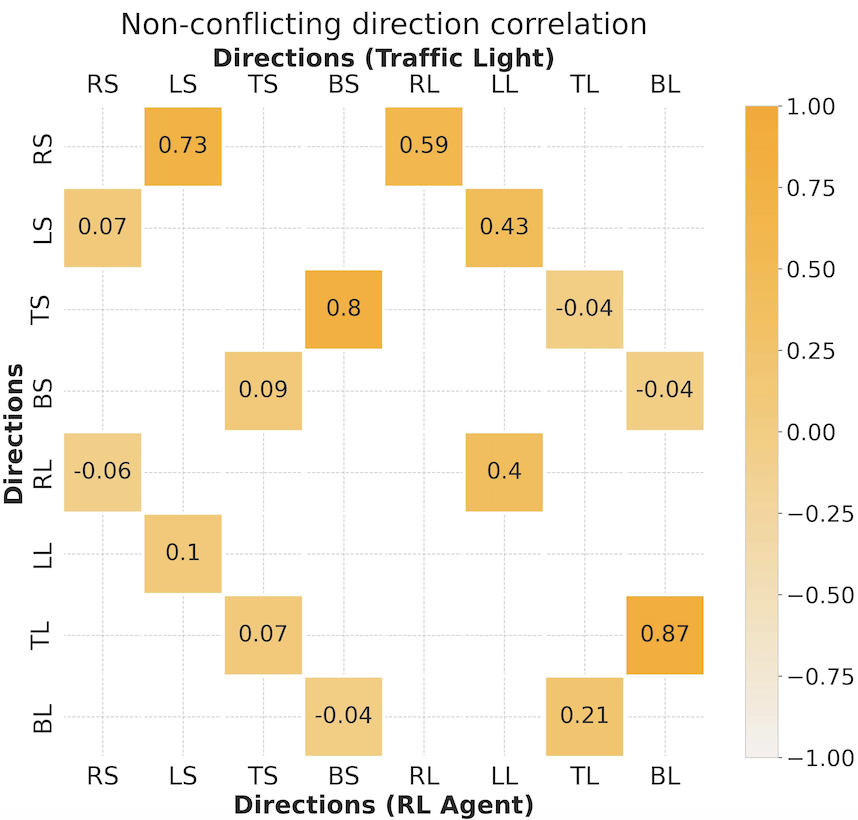}
   \caption{}
   \label{fig:outflow_non_conflict_499}
\end{subfigure}%
\begin{subfigure}{0.25\textwidth}
   \includegraphics[width=\linewidth]{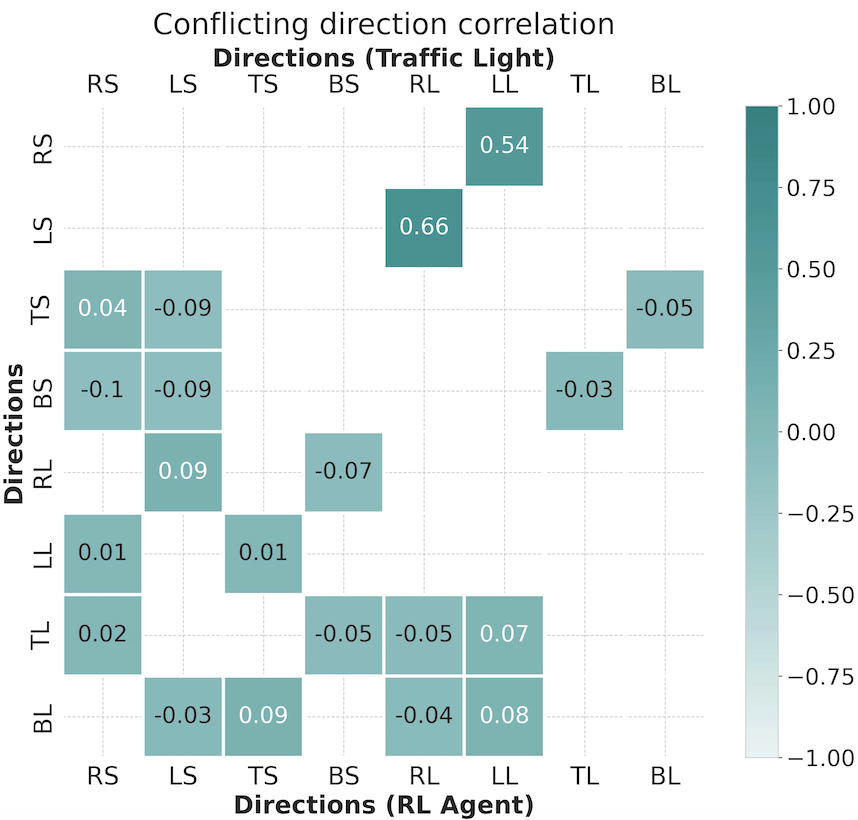}
   \caption{}
   \label{fig:outflow_conflict_499}
\end{subfigure}

\caption{PCC of queue length in (a) non-conflicting directions (b) conflicting directions for junction 499, outflow in (c) non-conflicting directions (d) conflicting directions for junction 499}
\label{fig:pcc_ql_outflow_499}
\end{figure*}
%%%%%%%%%%%%%%%%%%%%%%%%%%%%

\subsubsection{Queue Length Frequency Distribution}
%%%%%%%%%% Generated by ChatGPT (need to revised by me again)
Examining the distribution of queue lengths, a compelling observation comes to light: the queue lengths generated by Robot Vehicles (RVs) exhibit a frequent occurrence of shorter queues. This trend not only characterizes the RV-generated queue lengths but also echoes the same pattern observed in both non-conflicting and conflicting directions, as earlier revealed through our Pearson Coefficient Correlation (PCC) analysis.

This alignment in queue length patterns particularly stands out among non-conflicting directions, where a harmonious synchronization seems to manifest. It's worth noting that while this pattern holds true for the majority of scenarios, there are instances where we encounter limitations due to the unpredictability of traffic demand in specific directions. In such cases, a conclusive validation of this observation might be hindered by insufficient traffic flow data.

In stark contrast, the Traffic Lights (TL) system adheres to fixed-length traffic phases, lacking the dynamic adaptability to tailor traffic light phases based on fluctuating traffic demand scenarios for each direction. This rigidity sometimes leads to situations where, despite the absence of vehicles in a specific direction, traffic red lights impede progress, resulting in unwarranted delays—a suboptimal use of time and resources.

However, in the realm of Robot Vehicles, such inefficiencies are mitigated. RVs possess the innate capability to autonomously organize and optimize directional flows, intelligently considering the traffic situations in other directions. This agility ensures that time and resources are utilized more efficiently, paving the way for a smoother and more responsive traffic management system.

This observation proves we can get an idea of multimodality of RL agents during a stop-and-go scenario and management of RL agent's own
direction and coordinate with others. 

\subsubsection{Distribution of Platoon Sizes}
\begin{figure*}[htbp]
\centering
\begin{subfigure}{0.25\textwidth}
   \includegraphics[width=\linewidth]{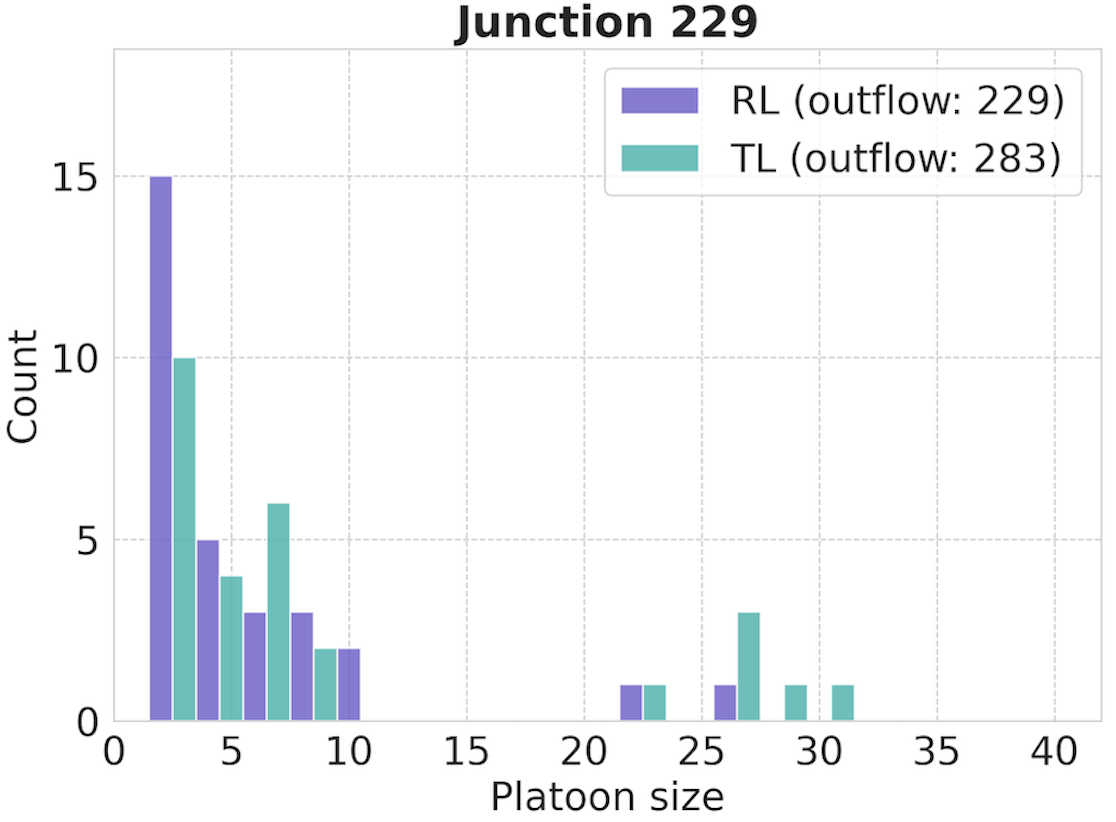}
   \caption{}
   \label{fig:platoon_size_229}
\end{subfigure}%
\begin{subfigure}{0.25\textwidth}
   \includegraphics[width=\linewidth]{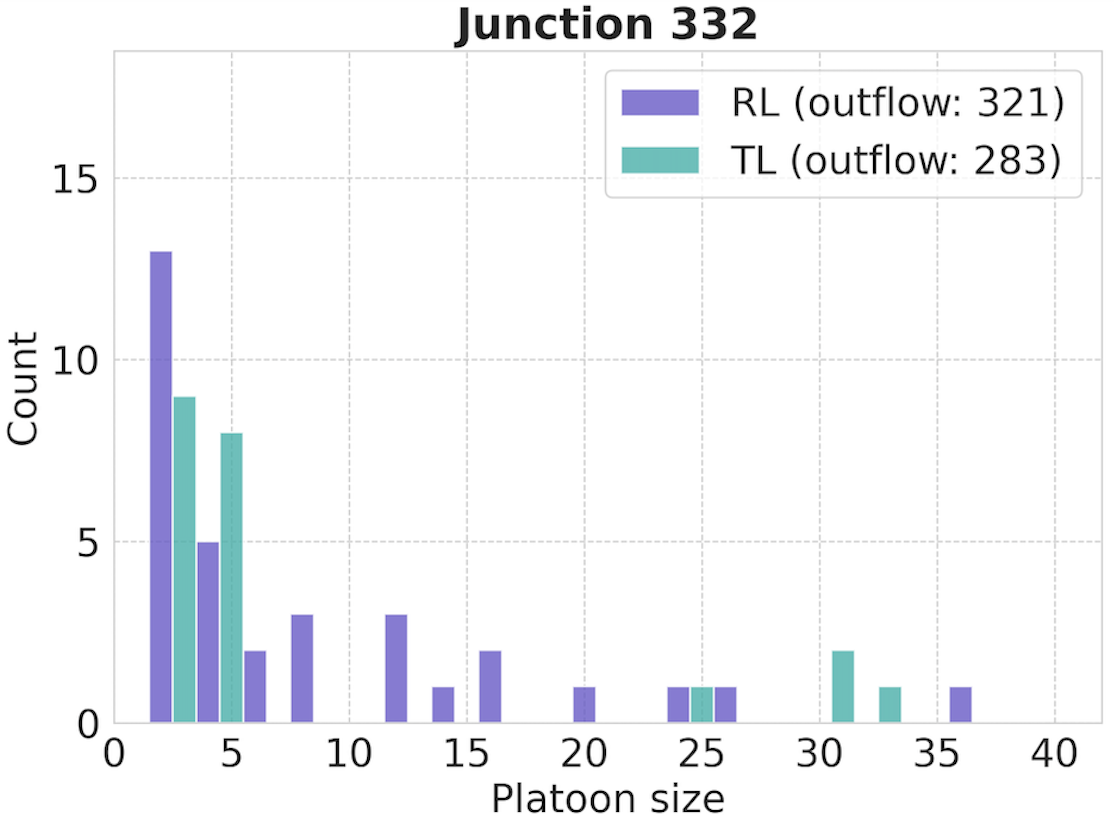}
   \caption{}
   \label{fig:platoon_size_332}
\end{subfigure}%
\begin{subfigure}{0.25\textwidth}
   \includegraphics[width=\linewidth]{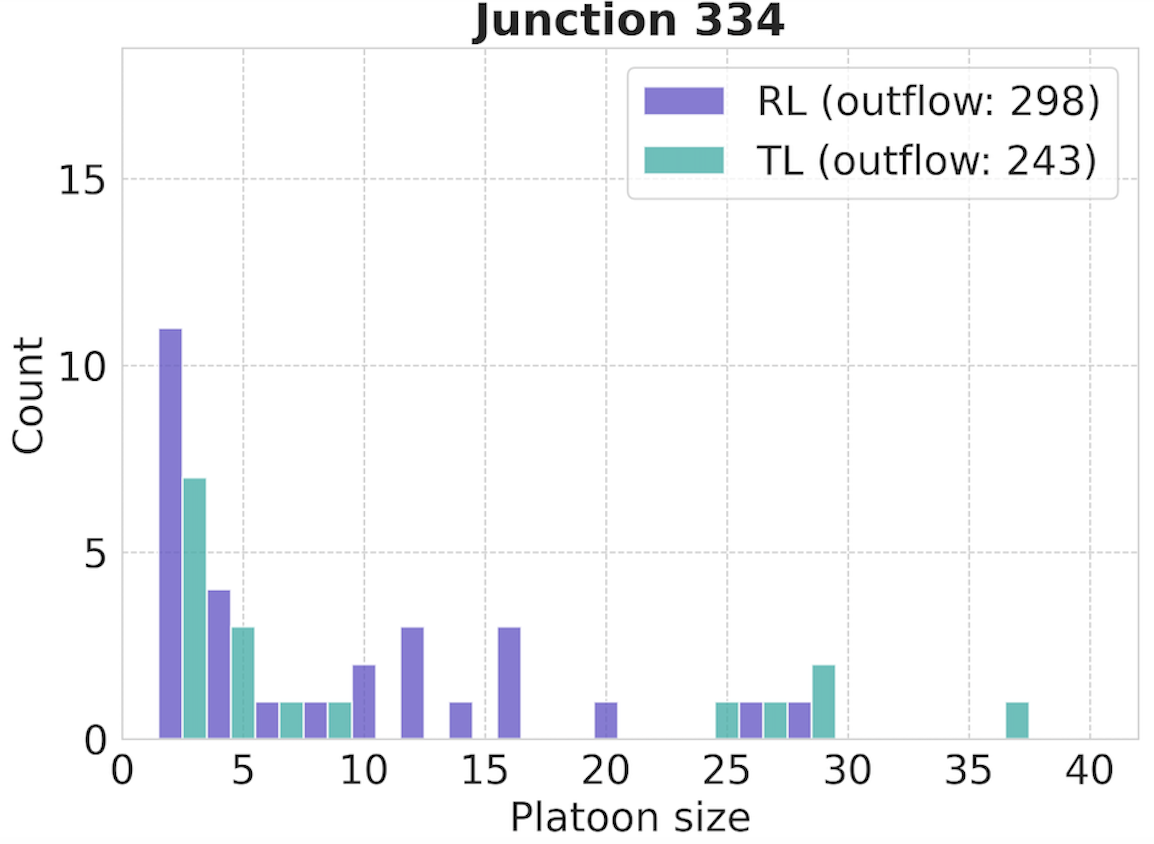}
   \caption{}
   \label{fig:platoon_size_334}
\end{subfigure}%
\begin{subfigure}{0.25\textwidth}
   \includegraphics[width=\linewidth]{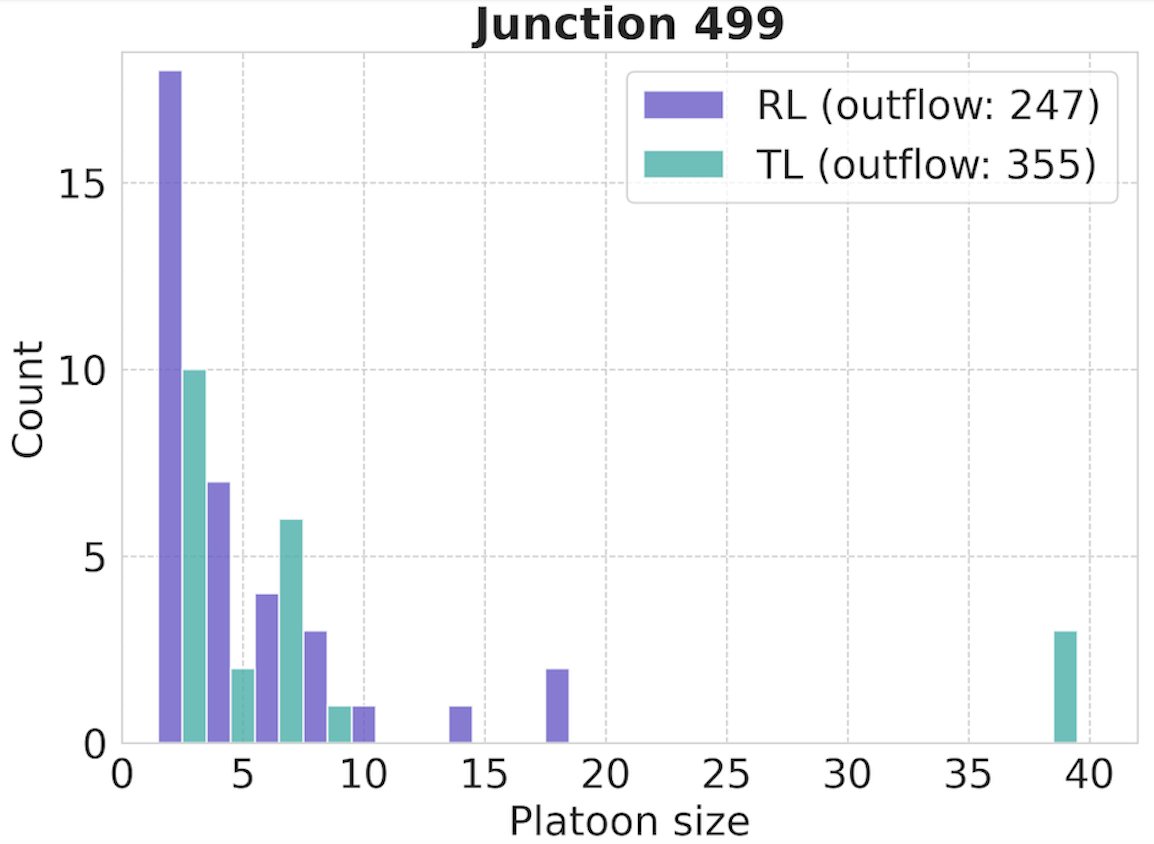}
   \caption{}
   \label{fig:platoon_size_499}
\end{subfigure}

\caption{Distribution of platoon size for junction (a) 229, (b) 332, (c) 334, (d) 499}
\label{fig:platoon_size}
\end{figure*}

A platoon refers to a group of vehicles that are closely following one another in a line or convoy, often at a relatively short distance and with little space between them. Platooning can occur naturally in traffic, especially on highways, where vehicles tend to travel at similar speeds and maintain a consistent following distance. A shorter size of the platoon promotes changes in the shorter queue and shorter wait times.  
Fig~\ref{fig:platoon_size} depicted the frequency distribution of platoon size for RL and TL. It demonstrated that RL creates shorter queues such as lengths 1, 2, and 3 more frequently compared to TL. Hence, it supports our hypothesis - 'RL acts as mini traffic light'. 

\section{Causal Analysis}
In this section, we discuss the investigation of RL agent's decision and its consequence on queue length and wait time. Many studies have conducted causal analysis regrading mobility~\cite{Wang2021Mobility,Lin2021Safety}. In this work, we have tried dowhy~\cite{dowhy}, and CausalImpact~\cite{causalimpact} tool for time series so far. For this purpose, we extracted a data set containing Time Step, Junction, Direction, Lane, Average Wait Time, Head of queue,  ql Type, and Action. \\
We followed two approaches for our causal investigation - (1) Causal effect of queue length of action in the same direction, (2) Causal effect of queue length of action in the same direction. 

\subsection{Dataset description for causality}
Table~\ref{tab:causality_dataset} provides a comprehensive summary of the instances associated with different directions that are used in the causality analysis. Each row represents a specific direction and the corresponding number of data instances such as queue length, action, wait time, etc.

\begin{table}[htbp]
\caption{List of instances in causality dataset}
\begin{center}
\begin{tabular}{c|c}
\hline
\hline
\textbf{Direction} & \textbf{No of instances}  \\
\hline
\hline
Topleft & 242 \\
\hline 
Rightstraight & 220 \\
\hline 
Bottomleft  & 211 \\
\hline 
Leftstraight & 192 \\
\hline
Topstraight  & 187  \\ 
\hline 
Bottomstraight & 61 \\
\hline 
Leftleft  & 61  \\
\hline 
Rightleft & No instance \\
\hline
\end{tabular}
\label{tab:causality_dataset}
\end{center}
\end{table}

\subsection{Causal Inference by CausalImpact}
\begin{figure}[htbp]
      \centering
      \includegraphics[scale=0.3, width=\linewidth]{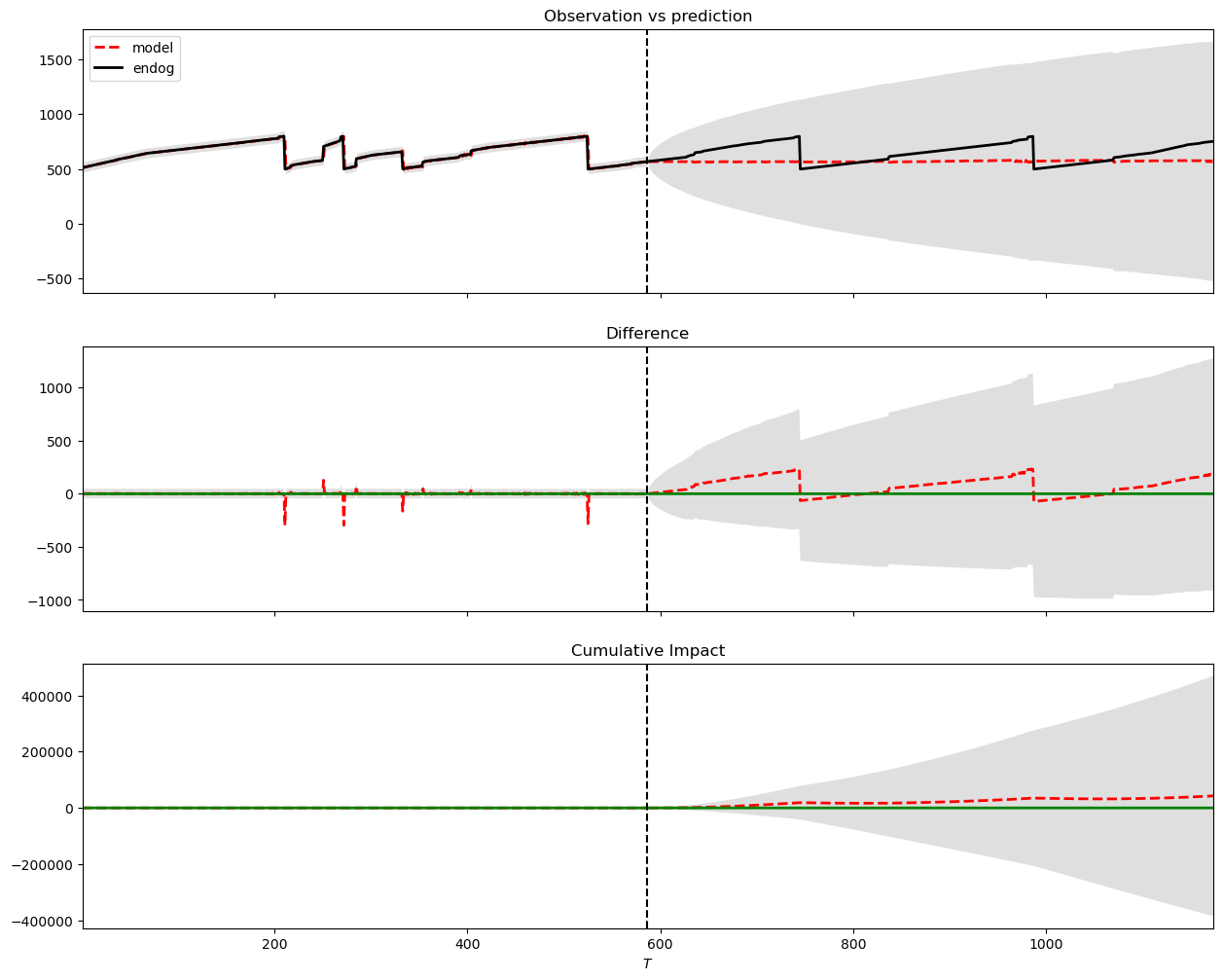}
      \caption{Illustration of the causal impact of queue length in all directions of 229}
      \label{ql_causal_impact_229}
   \end{figure}

In Fig~\ref{ql_causal_impact_229} the top plot typically shows the original queue length data, including during the pre-intervention period and the post-intervention period. CausalImpact estimates what would have happened to the outcome variable if the intervention hadn't occurred. This is called the "counterfactual" scenario. It uses Bayesian methods to generate uncertainty intervals. This plot shows the posterior predictive distribution of the outcome variable in the post-intervention period. This plot compares the observed data with the counterfactual prediction. It shows how the actual data deviates from what was expected in the absence of the intervention. The cumulative effects plot might display the cumulative impact of the intervention over time. The summary plot often combines key information, showing the observed data, predicted counterfactuals, and the impact estimate in a single graph.
We get p-value of 6.3\% and prob. of causal effect of 93.7\% for queue length in all directions of junction 229. However, CausalImpact tool can not find the cause of the causality. 

\subsection{Causal Inference by DoWhy}
The CausalImpact tool struggles to pinpoint the responsible factor behind a causal effect. In light of this, we have turned to another powerful causal inference tool called DoWhy. 

We have pursued two distinct approaches to delve into the causal effect of actions on queue length, specifically focusing on actions in the same direction shown in Table~\ref{tab:dowhy_same_dir} and those in different directions In Table~\ref{tab:dowhy_two_dir_2}. Our core assumption is that the action of the leading vehicle serves as the treatment, and we measure its impact on queue length in both the same and different directions. We have also explored the possibility of replacing queue length with average wait time in our analysis. While doing so, we observed that the changes in p-values and mean values were not remarkably significant.

However, it's worth noting that the results of our causality analysis lack definitive conclusions. This is primarily due to the varying flow in different directions and the fluctuating number of data instances, which can be attributed to the varying traffic demands.
  
\begin{table}[htbp]
\caption{List of P-value and Mean for same direction causality assumption}
\begin{center}
\begin{tabular}{c|c|c}
\hline
\hline
\textbf{Direction} & \textbf{P-value} & \textbf{Mean} \\
\hline
\hline
Bottomleft & $0.96$ & $-0.21$   \\
\hline
Bottomstraight & $0.94$ & $-0.89$ \\
\hline
Leftleft & $0.96$ & $1.39$  \\
\hline
Leftstraight & 0.9 & -0.94  \\
\hline
Rightstraight & 1.0 & 0.45 \\
\hline
Topleft  & 0.9 & -3.52 \\
\hline
Topstraight & 0.92 & -10.46  \\
\hline
Rightleft & no data & no data  \\
\hline
\end{tabular}
\label{tab:dowhy_same_dir}
\end{center}
\end{table}

% Expiremental comment
\begin{table}[htbp]
\caption{Causal effect for two different directions}
\begin{center}
\begin{tabular}{c|c|c|c}
\hline
\hline
\textbf{Direction 1} & \textbf{Direction 2} & \textbf{P-value} & \textbf{Mean}  \\
\hline
\hline
\multirow{6}{*}{Bottomleft} & Leftleft & 0.88 & 0.014 \\
\cline{2-3}
& Bottomstraight & 0.94 & 0.0 \\
\cline{2-3}
& Leftstraight & 0.86 & 0.153 \\
\cline{2-3}
& Rightstraight & 0.98 & 0.170 \\
\cline{2-3}
& Topleft & 0.96 & -0.413 \\
\cline{2-3}
& Topstraight & 0.96 & -0.063 \\
\hline
\multirow{6}{*}{Bottomstraight} & Leftleft & 0.94 & 0.0 \\
\cline{2-3}
& Bottomleft & 0.94 & 0.014 \\
\cline{2-3}
& Leftstraight & 0.94 & 0.066 \\
\cline{2-3}
& Rightstraight & 0.94 & 0.752 \\
\cline{2-3}
& Topleft & 0.58 & 8.882e-16 \\
\cline{2-3}
& Topstraight & 0.98 & 0.105 \\
\hline
\multirow{6}{*}{Leftstraight} & Leftleft & 0.86 & 0.153\\
\cline{2-3}
& Bottomleft & 0.86 & 0.153 \\
\cline{2-3}
& Bottomstraight & 0.94 & 0.066 \\
\cline{2-3}
& Rightstraight & 0.86 & 0.105 \\
\cline{2-3}
& Topleft & 0.86 & 0.122 \\
\cline{2-3}
& Topstraight & 0.86 & 0.142 \\
\hline
\multirow{6}{*}{Rightstraight} & Leftleft & 0.98 & 0.170 \\
\cline{2-3}
& Bottomleft & 0.98 & 0.170 \\
\cline{2-3}
& Bottomstraight & 0.98 & 0.170 \\
\cline{2-3}
& Leftstraight & 0.98 & 0.170 \\
\cline{2-3}
& Topleft & 0.98 & 0.170 \\
\cline{2-3}
& Topstraight & 0.98 & 0.170 \\
\hline
\multirow{6}{*}{Topleft} & Leftleft & 0.96 & -0.413 \\
\cline{2-3}
& Bottomleft & 0.96 & -0.413 \\
\cline{2-3}
& Bottomstraight & 0.96 & -0.413 \\
\cline{2-3}
& Leftstraight & 0.96 & -0.413 \\
\cline{2-3}
& Rightstraight & 0.96 & -0.413 \\
\cline{2-3}
& Topstraight & 0.96 & -0.068 \\
\hline
\multirow{6}{*}{Topstraight} & Leftleft & 0.96 & -0.063 \\
\cline{2-3}
& Bottomleft & 0.86 & -0.063 \\
\cline{2-3}
& Bottomstraight & 0.98 & 0.105 \\
\cline{2-3}
& Leftstraight & 0.86 & 0.142 \\
\cline{2-3}
& Rightstraight & 0.98 & 0.105 \\
\cline{2-3}
& Topleft & 0.96 & -0.068 \\
\hline
\end{tabular}
\label{tab:dowhy_two_dir_2}
\end{center}
\end{table}

\section{CONCLUSIONS}
\label{conclusion}
Our comprehensive analysis has provided valuable insights into the behavior of Reinforcement Learning (RL) agents and their interaction within complex traffic scenarios. We have addressed two fundamental research questions: the extent of multimodality in RL agent distributions during stop-and-go scenarios and the effective management of RVs in coordinating their movements. 

Our findings reveal that RL agents often exhibit multimodal distributions in queue length, outflow, and platoon size, indicating their adaptability and responsiveness in various traffic situations. Additionally, our examination of the Pearson coefficient correlation has illuminated the relationships between queue length and outflow, shedding light on how traffic dynamics are influenced by RL-controlled and Human-driven Vehicles (HVs). Notably, the PCC values are notably higher for non-conflicting directions in RL, demonstrating their capacity to synchronize and optimize traffic flow, thus mitigating congestion and reducing queue lengths.

Furthermore, our exploration of causal inference models has unveiled the factors influencing queue length in scenarios with different travel directions. This analysis offers critical insights into the decision-making processes of RL agents and their impact on traffic flow.

These findings provide valuable insights into how RL agents effectively manage traffic directions, answering part of our research question regarding RL's coordination abilities. Moreover, they support our hypothesis that RL functions as a dynamic "mini-traffic light," intelligently responding to traffic conditions rather than adhering rigidly to pre-programmed phases.

In conclusion, our comprehensive analysis not only enriches our understanding of RL agent behavior but also underscores their potential to revolutionize traffic management. The adaptability and responsiveness of RL agents hold promise for enhancing traffic flow and safety in complex traffic scenarios, ultimately contributing to the advancement of autonomous vehicle technology and more efficient traffic management systems.

There are many future research directions. 
First, we aim to analyze the RVs at the trajectory level~\cite{Li2019ADAPS,Shen2022IRL,Lin2022Attention}, under adversarial conditions~\cite{Poudel2021TrafficStateAttack,Shen2021Corruption,Villarreal2022AutoJoin}, and possibly with hardware~\cite{Poudel2022Micro}. This analysis intends to determine whether combining different elements would enhance the capabilities of robot vehicles further.
Secondly, we aim to scale up our study by incorporating intersectional mixed traffic into larger areas, ideally citywide road network. This will involve leveraging existing techniques for large-scale traffic simulation, reconstruction, and prediction~\cite{Wilkie2015Virtual,Li2017CityFlowRecon,Li2017CitySparseITSM,Li2018CityEstIET,Lin2019Compress,Lin2019BikeTRB,Chao2020Survey,Lin2022GCGRNN}. 
Third, we want to analyze the impact of various vehicle networks via network optimization~\cite{Wickman2022SparRL}, and heterogeneous traffic via multi-agent learning and optimization~\cite{Wickman2023Species,wickman2021lrn}. 
Lastly, we want to combine mixed traffic modeling with crowd simulation and virtual humans~\cite{Li2013Memory,Durupinar2016Individual,Li2012Commonsense,Li2012Apprentice,Li2012Distribution,Li2011Purpose} so that together we can model more realistic road uses with pedestrians as intelligent virtual characters.

% \clearpage
% \input{sections/appendix}

%%%%%%%%%%%%%%%%%%%%%%%%%%%%%%%%%%%%%%%%%%%%%%%%%%%%%%%%%%%%%%%%%%%%%%%%%%%%%%%%

%\section*{APPENDIX}


% \section*{ACKNOWLEDGMENT}

% The preferred spelling of the word ÒacknowledgmentÓ in America is without an ÒeÓ after the ÒgÓ. Avoid the stilted expression, ÒOne of us (R. B. G.) thanks . . .Ó  Instead, try ÒR. B. G. thanksÓ. Put sponsor acknowledgments in the unnumbered footnote on the first page.

\bibliographystyle{unsrt}
\bibliography{ref}

\begin{thebibliography}{10}

\bibitem{waymo}
Waymo - self-driving technology company.
\newblock \url{https://waymo.com}.
\newblock Accessed on September 5, 2023.

\bibitem{waymo2022citybybay}
Waymo.
\newblock Taking our next step in the city by the bay.
\newblock \url{https://blog.waymo.com/2022/03/taking-our-next-step-in-city-by-bay.html}.

\bibitem{cruise}
{Cruise Automation, Inc.}
\newblock Cruise - self-driving cars, Year of Access.
\newblock Accessed on Month Day, Year.

\bibitem{cruise2023onemillion}
Cruise Automation.
\newblock One million driverless miles.
\newblock \url{https://getcruise.com/news/blog/2023/one-million-driverless-miles/}.

\bibitem{nhtsa}
National Highway Traffic~Safety Administration.
\newblock Nhtsa.
\newblock \url{https://www.nhtsa.gov}.

\bibitem{villarreal2023can}
Michael Villarreal, Bibek Poudel, and Weizi Li.
\newblock Can chatgpt enable its? the case of mixed traffic control via reinforcement learning.
\newblock In {\em IEEE International Conference on Intelligent Transportation Systems (ITSC)}, 2023.

\bibitem{villarreal2023hybrid}
Michael Villarreal, Bibek Poudel, Jia Pan, and Weizi Li.
\newblock Hybrid traffic control and coordination from pixels.
\newblock {\em arXiv preprint arXiv:2302.09167}, 2023.

\bibitem{wang2023learning}
Dawei Wang, Weizi Li, Lei Zhu, and Jia Pan.
\newblock Learning to control and coordinate hybrid traffic through robot vehicles at complex and unsignalized intersections.
\newblock {\em arXiv preprint arXiv:2301.05294}, 2023.

\bibitem{Wang2023Privacy}
Dawei Wang, Weizi Li, and Jia Pan.
\newblock Large-scale mixed traffic control using dynamic vehicle routing and privacy-preserving crowdsourcing.
\newblock {\em IEEE Internet of Things Journal}, 2023.

\bibitem{dresner2008multiagent}
Kurt Dresner and Peter Stone.
\newblock A multiagent approach to autonomous intersection management.
\newblock {\em Journal of artificial intelligence research}, 31:591--656, 2008.

\bibitem{fajardo2011automated}
David Fajardo, Tsz-Chiu Au, S~Travis Waller, Peter Stone, and David Yang.
\newblock Automated intersection control: Performance of future innovation versus current traffic signal control.
\newblock {\em Transportation Research Record}, 2259(1):223--232, 2011.

\bibitem{sharon2017protocol}
Guni Sharon and Peter Stone.
\newblock A protocol for mixed autonomous and human-operated vehicles at intersections.
\newblock In {\em Autonomous Agents and Multiagent Systems: AAMAS 2017 Workshops, Best Papers, S{\~a}o Paulo, Brazil, May 8-12, 2017, Revised Selected Papers 16}, pages 151--167. Springer, 2017.

\bibitem{jang2019simulation}
Kathy Jang, Eugene Vinitsky, Behdad Chalaki, Ben Remer, Logan Beaver, Andreas~A Malikopoulos, and Alexandre Bayen.
\newblock Simulation to scaled city: zero-shot policy transfer for traffic control via autonomous vehicles.
\newblock In {\em Proceedings of the 10th ACM/IEEE International Conference on Cyber-Physical Systems}, pages 291--300, 2019.

\bibitem{yang2020intelligent}
Hao Yang and Ken Oguchi.
\newblock Intelligent vehicle control at signal-free intersection under mixed connected environment.
\newblock {\em IET Intelligent Transport Systems}, 14(2):82--90, 2020.

\bibitem{Wang2021Mobility}
Songhe Wang, Kangda Wei, Lei Lin, and Weizi Li.
\newblock Spatial-temporal analysis of {COVID}-19's impact on human mobility: the case of the united states.
\newblock In {\em The 20th and 21st Joint COTA International Conference of Transportation Professionals}, 2021.

\bibitem{Lin2021Safety}
Lei Lin, Feng Shi, and Weizi Li.
\newblock Assessing inequality, irregularity, and severity regarding road traffic safety during covid-19.
\newblock {\em Scientific Reports}, 11(13147):1--7, 2021.

\bibitem{dowhy}
Jasjeet~S. Sekhon, Brenton Kenkel, and Philip VanderWeele.
\newblock Dowhy: A python library for causal inference, 2023.

\bibitem{causalimpact}
{Google, Inc.}
\newblock {\em CausalImpact: A Package for Causal Inference Using Bayesian Structural Time-Series Models}, Year of Access.
\newblock Accessed on Date.

\bibitem{Li2019ADAPS}
Weizi Li, David Wolinski, and Ming~C. Lin.
\newblock {ADAPS}: Autonomous driving via principled simulations.
\newblock In {\em IEEE International Conference on Robotics and Automation (ICRA)}, pages 7625--7631, 2019.

\bibitem{Shen2022IRL}
Yu~Shen, Weizi Li, and Ming~C. Lin.
\newblock Inverse reinforcement learning with hybrid-weight trust-region optimization and curriculum learning for autonomous maneuvering.
\newblock In {\em IEEE/RSJ International Conference on Intelligent Robots and Systems (IROS)}, pages 7421--7428, 2022.

\bibitem{Lin2022Attention}
Lei Lin, Weizi Li, Huikun Bi, and Lingqiao Qin.
\newblock Vehicle trajectory prediction using {LSTM}s with spatial-temporal attention mechanisms.
\newblock {\em IEEE Intelligent Transportation Systems Magazine}, 14(2):197–208, 2022.

\bibitem{Poudel2021TrafficStateAttack}
Bibek Poudel and Weizi Li.
\newblock Black-box adversarial attacks on network-wide multi-step traffic state prediction models.
\newblock In {\em IEEE International Conference on Intelligent Transportation Systems (ITSC)}, pages 3652--3658, 2021.

\bibitem{Shen2021Corruption}
Yu~Shen, Laura Zheng, Manli Shu, Weizi Li, Tom Goldstein, and Ming~C. Lin.
\newblock Gradient-free adversarial training against image corruption for learning-based steering.
\newblock In {\em Advances in Neural Information Processing Systems (NeurIPS)}, pages 26250--26263, 2021.

\bibitem{Villarreal2022AutoJoin}
Michael Villarreal, Bibek Poudel, Ryan Wickman, Yu~Shen, and Weizi Li.
\newblock Autojoin: Efficient adversarial training for robust maneuvering via denoising autoencoder and joint learning.
\newblock 2023.

\bibitem{Poudel2022Micro}
Bibek Poudel, Thomas Watson, and Weizi Li.
\newblock Learning to control dc motor for micromobility in real time with reinforcement learning.
\newblock In {\em IEEE International Conference on Intelligent Transportation Systems (ITSC)}, pages 1248--1254, 2022.

\bibitem{Wilkie2015Virtual}
David Wilkie, Jason Sewall, Weizi Li, and Ming~C. Lin.
\newblock Virtualized traffic at metropolitan scales.
\newblock {\em Frontiers in Robotics and AI}, 2:11, 2015.

\bibitem{Li2017CityFlowRecon}
Weizi Li, David Wolinski, and Ming~C. Lin.
\newblock City-scale traffic animation using statistical learning and metamodel-based optimization.
\newblock {\em ACM Trans. Graph.}, 36(6):200:1--200:12, 2017.

\bibitem{Li2017CitySparseITSM}
Weizi Li, Dong Nie, David Wilkie, and Ming~C. Lin.
\newblock Citywide estimation of traffic dynamics via sparse {GPS} traces.
\newblock {\em IEEE Intelligent Transportation Systems Magazine}, 9(3):100--113, 2017.

\bibitem{Li2018CityEstIET}
Weizi Li, Meilei Jiang, Yaoyu Chen, and Ming~C. Lin.
\newblock Estimating urban traffic states using iterative refinement and wardrop equilibria.
\newblock {\em IET Intelligent Transport Systems}, 12(8):875--883, 2018.

\bibitem{Lin2019Compress}
Lei Lin, Weizi Li, and Srinivas Peeta.
\newblock Efficient data collection and accurate travel time estimation in a connected vehicle environment via real-time compressive sensing.
\newblock {\em Journal of Big Data Analytics in Transportation}, 1(2):95--107, 2019.

\bibitem{Lin2019BikeTRB}
Lei Lin, Weizi Li, and Srinivas Peeta.
\newblock Predicting station-level bike-sharing demands using graph convolutional neural network.
\newblock In {\em Transportation Research Board 98th Annual Meeting (TRB)}, 2019.

\bibitem{Chao2020Survey}
Qianwen Chao, Huikun Bi, Weizi Li, Tianlu Mao, Zhaoqi Wang, Ming~C. Lin, and Zhigang Deng.
\newblock A survey on visual traffic simulation: Models, evaluations, and applications in autonomous driving.
\newblock {\em Computer Graphics Forum}, 39(1):287--308, 2020.

\bibitem{Lin2022GCGRNN}
Lei Lin, Weizi Li, and Lei Zhu.
\newblock Data-driven graph filter based graph convolutional neural network approach for network-level multi-step traffic prediction.
\newblock {\em Sustainability}, 14(24):16701, 2022.

\bibitem{Wickman2022SparRL}
Ryan Wickman, Xiaofei Zhang, and Weizi Li.
\newblock A generic graph sparsification framework using deep reinforcement learning.
\newblock In {\em IEEE International Conference on Data Mining (ICDM)}, pages 1221--1226, 2022.

\bibitem{Wickman2023Species}
Ryan Wickman, Bibek Poudel, Michael Villarreal, Xiaofei Zhang, and Weizi Li.
\newblock Efficient quality-diversity optimization through diverse quality species.
\newblock In {\em Genetic and Evolutionary Computation Conference (GECCO)}, 2023.

\bibitem{wickman2021lrn}
Ryan Wickman, Xiaofei Zhang, and Weizi Li.
\newblock Lrn: Limitless routing networks for effective multi-task learning.
\newblock 2021.

\bibitem{Li2013Memory}
Weizi Li, Tim Balint, and Jan~M. Allbeck.
\newblock Using a parameterized memory model to modulate npc {AI}.
\newblock In {\em Proceedings of the 13th International Conference on Intelligent Virtual Agents (IVA)}, pages 1--14, 2013.

\bibitem{Durupinar2016Individual}
Funda Durupinar, Nuria Pelechano, Jan Allbeck, Norman Badler, and Weizi Li.
\newblock {\em Individual Differences}, pages 159--171.
\newblock Springer International Publishing, 2016.

\bibitem{Li2012Commonsense}
Weizi Li and Jan~M. Allbeck.
\newblock Virtual humans: Evolving with common sense.
\newblock In {\em Proceedings of the 5th International Conference on Motion in Games (MIG)}, pages 182--193, 2012.

\bibitem{Li2012Apprentice}
Weizi Li and Jan~M. Allbeck.
\newblock The virtual apprentice.
\newblock In {\em Proceedings of the 12th International Conference on Intelligent Virtual Agents (IVA)}, pages 15--27, 2012.

\bibitem{Li2012Distribution}
Weizi Li, Zichao Di, and Jan~M. Allbeck.
\newblock Crowd distribution and location preference.
\newblock {\em Computer Animation and Virtual Worlds}, 23(3-4):343--351, 2012.

\bibitem{Li2011Purpose}
Weizi Li and Jan~M. Allbeck.
\newblock Populations with purpose.
\newblock In {\em Proceedings of the 4th International Conference on Motion in Games (MIG)}, pages 132--143, 2011.

\end{thebibliography}

\end{document}